\documentclass[a4paper,fleqn]{cas-dc}

\usepackage[numbers]{natbib}

\def\tsc#1{\csdef{#1}{\textsc{\lowercase{#1}}\xspace}}
\tsc{WGM}
\tsc{QE}
\tsc{EP}
\tsc{PMS}
\tsc{BEC}
\tsc{DE}
\begin{document}
\let\WriteBookmarks\relax
\def\floatpagepagefraction{1}
\def\textpagefraction{.001}
\shorttitle{Urban Scale Surface Reconstruction }
\shortauthors{Z.X. Li et~al.}

\title [mode = title]{ULSR-GS: Urban Large-scale Surface Reconstruction Gaussian Splatting with Multi-View Geometric Consistency}

% \tnotetext[1]{This document is the results of the research
%    project funded by the National Science Foundation.}

% \tnotetext[2]{The second title footnote which is a longer text matter
%    to fill through the whole text width and overflow into
%    another line in the footnotes area of the first page.}

\author[inst1,inst5]{Zhuoxiao Li \dag}
\author[inst1,inst3]{Shanliang Yao\dag}
\author[inst1]{Taoyu Wu}
\author[inst1,]{Yong Yue}
\author[inst2]{Wufan Zhao}
\author[inst6]{Rongjun Qin}
\author[inst4,inst5]{\'{A}ngel F. Garc\'{i}a-Fern\'{a}ndez}
\author[inst5,inst7]{Andrew Levers}
\author[inst5]{Jason Ralph}

\author[inst1]{Xiaohui Zhu \corref{cor1}}

\affiliation[inst1]{
            organization={School of Advanced Technology},%Department and Organization
            addressline={ Xi'an Jiaotong-Liverpool University}, 
            city={Suzhou},
            postcode={215123}, 
            country={China}
            }
\affiliation[inst2]{
            organization={Urban Governance and Design Thrust},%Department and Organization
            addressline={The Hong Kong University of Science and Technology (Guangzhou)}, 
            city={Guangzhou},
            postcode={511453}, 
            country={China}
            }
\affiliation[inst5]{
            organization={Department of Electrical Engineering and Electrnics},%Department and Organization
            addressline={University of Liverpool}, 
            city={Liverpool},
            postcode={L69 3GJ}, 
            country={UK}
            } 
\affiliation[inst6]{
            organization={Department of Civil,Environmental and Geodetic Engineering},%Department and Organization
            addressline={The Ohio State University}, 
            city={Columbus},
            postcode={OH 43210}, 
            country={USA}
            }

\affiliation[inst3]{
            organization={School of Information Engineering},%Department and Organization
            addressline={Yancheng Institute Technology}, 
            city={Yancheng},
            postcode={224051}, 
            country={China}
            }

\affiliation[inst4]{
            organization={IPTC, ETSI de Telecomunicaci\'on},%Department and Organization
            addressline={Universidad Polit\'ecnica de Madrid}, 
            city={Madrid},
            postcode={28040}, 
            country={Spain}
            }

\affiliation[inst7]{
            organization={Institute for Digital Engineering and Autonomous Systems},%Department and Organization
            addressline={University of Liverpool}, 
            city={Liverpool},
            postcode={L69 3GJ}, 
            country={UK}
            } 

\cortext[cor1]{Corresponding author: Xiaohui.Zhu@xjtlu.edu.cn} 
\cortext[cor2]{\dag Equal Contribution}
% \cortext[cor2]{Principal corresponding author}

\begin{abstract}
Recent advances in 2D Gaussian Splatting (2DGS) have demonstrated compelling rendering efficiency and mesh extraction capabilities. However, its application to large-scale aerial photogrammetry, especially using oblique UAV imagery, remains limited due to three primary challenges:
(1) suboptimal image selection in scene partitioning strategies failing to scale effectively;
(2) densification pipelines that rely primarily on single-view constraints resulting in over-smoothed reconstructions and loss of fine geometric detail; and
(3) the absence of multi-view geometric consistency constraints leading to surface artifacts and inconsistencies.
To address these limitations, we propose ULSR-GS, a novel method tailored for high-resolution surface reconstruction in urban-scale environments.
Firstly, we propose a point-to-photo partitioning strategy that segments the scene based on the sparse SfM point cloud and assigns only the most relevant images to each sub-region, which resolves key scalability bottlenecks.
Secondly, we propose a multi-view guided densification strategy that enforces adaptive geometric consistency across views, overcoming the limitations of single-view-based densifications.
Lastly, we introduce consistency-aware loss functions that explicitly regulate depth and normal alignment across views, significantly enhancing surface fidelity.
Extensive experiments on large-scale aerial benchmark datasets demonstrate that ULSR-GS consistently outperforms existing single- and multi-GPU Gaussian Splatting methods. Furthermore, compared to MVS pipelines, our approach achieves comparable or superior geometric quality while being substantially more time-efficient, making it a practical solution for scalable 3D modeling in digital twin and urban mapping applications. Project page: \href{https://ulsrgs.github.io}{https://ulsrgs.github.io}.
\end{abstract}


\maketitle

\section{Introduction}\label{sec:intro}
High-fidelity 3D surface reconstruction at urban scale remains a longstanding challenge in photogrammetry and computer graphics, with critical applications in urban planning, simulation, and digital twins \cite{wu2024integrating, pan2024deep,ccoltekin2020extended}. Traditionally, urban-scale surface reconstruction from aerial oblique imagery has relied heavily on Multi-View Stereo (MVS) techniques \cite{schoenberger2016colmap,li2020novel,xu2022acmmp_mvs,xu2020planar_mvs, bleyer2011patchmatch, YANG2015262isprs1,ZHU201547isprs2, PALMA2018328isprs3,LIU2024361isprs4,YANG2022302isprs5,GAO2023446isprs6,LIU202342isprs7,ZHANG202427isprs8,LIAO2024173isprs9}, which reconstruct dense point clouds from image correspondences and subsequently extract meshes through surface reconstruction algorithms, such as Poisson Surface Reconstruction \cite{kazhdan2006poisson, kazhdan2013screened} or Delaunay triangulation-based methods \cite{Delaunay}. While effective, these traditional pipelines often encounter computational inefficiencies when scaling to urban-scale datasets comprising thousands of high-resolution images. Recent advancements in novel view  synthesis (NVS) have led to from Neural Radiance Fields (NeRFs) \cite{xu2022pointnerf,zhang2020nerf++,cao2022mobilnerf,barron2021mipnerf,barron2022mipnerf360} to the emergence of 3D Gaussian Splatting (3DGS) \cite{3dgs}, a breakthrough approach enabling fast photorealistic rendering quality with unprecedented efficiency by directly optimizing sparse 3D Gaussian primitives. In addition to achieving compelling visual realism, recent studies have demonstrated that 3DGS and its variants can serve as effective tools for extracting geometry-rich meshes \cite{huang20242dgs,chen2024pgsr,li2024mvgsplatting}. 

Scaling Gaussian Splatting (GS) to city-scale surface reconstruction tasks is both an engineering and algorithmic challenge. Issue (1): The vanilla GS training pipeline \cite{3dgs,huang20242dgs,yu2024mipsplatting,Yu2024GOF} does not inherently support multi-GPU parallel processing. For instance, even high-performance GPUs like the RTX 4090 (24GB VRAM) can manage only approximately 8 million primitives, limiting applications to relatively small datasets such as the Mip-NeRF 360 dataset \cite{barron2022mipnerf360}. Consequently, current large-scale GS approaches predominantly rely on a divide-and-conquer strategy by
partitioning extensive scenes into smaller sub-scenes based on camera projections or clustering of camera poses obtained from Structure-from-Motion (SfM), subsequently distributing these partitions across multiple GPUs \cite{vastgaussian, hierarchicalgaussians24, chen2024dogaussiandistributedorientedgaussiansplatting, liu2024citygaussian, liu2024citygaussianv2efficientgeometricallyaccurate}. While effective for controlled, uniform camera distributions such as five-directional aerial surveys \cite{UrbanScene3D, xiong2024gauuscene}, these image-based methods struggle with close-range oblique photogrammetry scenarios, where irregular and complex camera trajectories result in uneven or inadequate training data distribution. Issue (2): Existing approaches necessitate merging sub-scenes before mesh extraction, a process requiring substantial computational resources. For example, methods employing TSDF fusion with a 0.2-meter voxel size consume over 300GB RAM for reconstructing merely 1,000 images, making scalability to city-scale scenes with tens of thousands of images impractical. Additionally, extracting meshes separately introduces further complications, such as determining optimal sub-scene boundaries, often leading to unnecessary computational overhead and fragmented meshes. Issue (3): Another significant challenge in GS-based large-scale reconstruction is the densification problem. GS explicitly represents scenes using discrete Gaussian primitives, meaning the richer and denser the primitives, the more accurately high-frequency geometric details can be captured \cite{yu2024mipsplatting,chen2024mvsplat,li2024mvgsplatting,fan2024trim}. However, GS often struggles with accurately reconstructing low-frequency or textureless regions, resulting in incomplete or overly smooth surfaces \cite{huang20242dgs,fan2024trim,chen2024pgsr}. Issue (4): Existing GS-based methods predominantly utilize single-view geometric consistency constraints during training \cite{yu2024mipsplatting,chen2024mvsplat,li2024mvgsplatting,fan2024trim}, which fail to leverage comprehensive multi-view information. This single-view approach can cause inconsistencies and inaccuracies, such as floaters or misaligned primitives, particularly in complex urban scenarios.

To address these critical issues, we propose ULSR-GS. Our key insight is combining detail-preserving optimization within each divided sub-region and enforcing multi-view consistency during training, which makes it possible for 2DGS to independently reconstruct each tile with high geometric fidelity. These sub-regions can then be seamlessly integrated into a globally consistent urban-scale surface without additional fusion or post-processing steps.  Unlike other methods that partition scenes solely by camera positions \cite{liu2024citygaussian,yuchen2024dogaussian,vastgaussian,liu2024citygaussianv2efficientgeometricallyaccurate}, we propose a point-to-photo partitioning strategy driven by the spatial distribution of sparse SfM points and their associated views. For each candidate sub-region, we meticulously select an optimal subset of images for each SfM point, ensuring comprehensive and relevant coverage.
We then introduce a multi-view constrained densification approach within the optimization process for each partitioned sub-region. This strategy explicitly enforces geometric consistency across multiple views by penalizing discrepancies in depth and normal orientations of corresponding points across different images. Consequently, our multi-view densification preserves essential fine details, which are preserved when sub-regions are optimized independently. Collectively, the innovative integration of our point-to-photo partitioning and multi-view densification strategies positions ULSR-GS as a highly effective and scalable solution for accurate and detailed large-scale urban surface reconstructions.

Our contributions are summarized as follows:

\begin{itemize}
    \item We present ULSR-GS, a novel method specifically designed to overcome the limitations of existing GS-based works in large-scale surface reconstruction tasks.
    
    \item  A  partitioning strategy that supports extracting the geometry of each trained GS sub-region separately, without the need for merging  and then extracting.
    
    \item An densification method based on multi-view depth aggregation, which can improve rendering and geometry reconstruction quality. 
    
    \item Two efficient multi-view consistency constraints for generating detailed and accurate reconstructions in large-scale urban environments.
\end{itemize}

\section{Related work}

\subsection{Large-Scale Novel View Synthesis}
Recent developments in novel view synthesis \cite{barron2021mipnerf,chen2022gpnerf_mvs, zhang2020nerf++, mildenhall2021nerf} include several innovative approaches tailored for large-scale environments \cite{xiangli2022bungeenerf, tancik2022blocknerf, mi2023switchnerf, turki2022mega}. BungeeNeRF first addresses the challenge of rendering city-scale scenes with drastic scale variations by introducing a progressive neural radiance field that adapts to different levels of detail \citep{xiangli2022bungeenerf}. BlockNeRF segments a city into distinct blocks and strategically allocates training views based on their spatial locations \cite{tancik2022blocknerf}. Mega-NeRF takes a grid-based approach, assigning each pixel in an image to various grids corresponding to the rays it traverses \cite{turki2022mega}. In contrast to these partitioning methods, Switch-NeRF employs a mixture-of-experts framework that facilitates scene decomposition, allowing for more nuanced representation \cite{mi2023switchnerf}. Grid-NeRF, while not focusing on scene decomposition, combines NeRF-based techniques with grid-based methodologies to enhance rendering efficiency \cite{xu2023gridguided}. 

Moreover, the adoption of 3D Gaussian Splatting \cite{3dgs} has emerged as a promising approach, improving both reconstruction accuracy and rendering speed \cite{yu2024mipsplatting, lu2024scaffold, cheng2024gaussianpro}.
GS-based large-scale rendering methods primarily handle large-scale scene rendering through camera-position-based partitioning, followed by point cloud selection or segmentation \cite{vastgaussian,yuchen2024dogaussian,liu2024citygaussian,hierarchicalgaussians24,chen2024gigags}. This strategy ensures that training images for each region remain as similar as possible, enabling efficient techniques like Level of Detail (LOD) to accelerate rendering \cite{liu2024citygaussian, hierarchicalgaussians24, chen2024gigags}. This partitioning approach is particularly effective for terrestrial sequences \cite{hierarchicalgaussians24} and aerial five-directional oblique photography \cite{vastgaussian, yuchen2024dogaussian, chen2024gigags}. However, when dealing with scenarios involving specific path planning algorithms, these partitioning methods may struggle to adapt efficiently, leading to suboptimal training data distribution and reduced reconstruction quality for dynamically changing viewpoints.

\subsection{3D Surface Reconstruction}

3D reconstruction has evolved from traditional Multi-View Stereo (MVS) \cite{schoenberger2016colmap,li2020novel,xu2022acmmp_mvs,xu2020planar_mvs, bleyer2011patchmatch, YANG2015262isprs1,ZHU201547isprs2, PALMA2018328isprs3,LIU2024361isprs4,YANG2022302isprs5,GAO2023446isprs6,LIU202342isprs7,ZHANG202427isprs8,LIAO2024173isprs9} to advanced neural methods \cite{wang2021neus,oechsle2021unisurf_neus, yariv2023bakedsdf}, and mesh extraction methods transform implicit representations into explicit 3D models. Poisson Surface-based methods extract surfaces from point clouds but often produce overly smooth results, lacking fine details \cite{guedon2023sugar}. Marching Cubes converts volumetric data into polygonal meshes, but its grid-based nature can lead to artifacts and increased memory usage \cite{lorensen1998marching, li2024mvgsplatting}. Neural implicit representations encode 3D geometry continuously, enabling smooth modeling and handling complex topologies \cite{wang2021neus, yariv2023bakedsdf, yariv2021volume}. However, these methods require significant computational resources and often struggle with fine geometric details. Mesh extraction from these representations also suffers from discretization artifacts. Neural Radiance Fields (NeRF) use volumetric rendering to achieve high-quality view synthesis \cite{mildenhall2021nerf}. Despite their impressive results, NeRF-based methods are computationally intensive, making them less practical for real-time applications \cite{barron2021mipnerf, turki2022mega}. Recent works extract explicit geometry from NeRF, but converting density fields to surfaces in large-scale scenes remains challenging and may produce noisy results \cite{li2023neuralangelo}. 

Some research has begun to explore mesh extraction using 3DGS \cite{3dgs}. Early work combined 3DGS with Poisson Surface Reconstruction to achieve reconstruction \cite{guedon2023sugar}. Subsequently, studies used rendered depth maps with Truncated Signed Distance Function (TSDF) fusion to reconstruct surfaces \cite{werner2014tsdf},   \cite{turkulainen2024dnsplatter}, improving accuracy by modifying the depth calculation method \cite{huang20242dgs,zhang2024rade,chen2024pgsr,li2024mvgsplatting}, and optimizing the densification process during training \cite{fan2024trim,li2024mvgsplatting}. Additionally, a study utilized Gaussian opacity fields combined with a marching tetrahedra approach to achieve high-quality reconstruction results \cite{Yu2024GOF}. Existing GS-based methods perform inadequately in large-scale scenes, struggling to accurately reconstruct intricate geometric details and prone to generating significant artifacts. Using the partition rendering methods \cite{vastgaussian,liu2024citygaussian,li2023matrixcity} to split the scene into different sub-regions can solve this problem well, and each sub-region can obtain sufficient densification.  However, the current partitioning methods lack optimization of the mesh extraction task and usually require mesh extraction after merging the entire scene.

\begin{figure*}[!ht]
    \centering
    \includegraphics[width=1\linewidth]{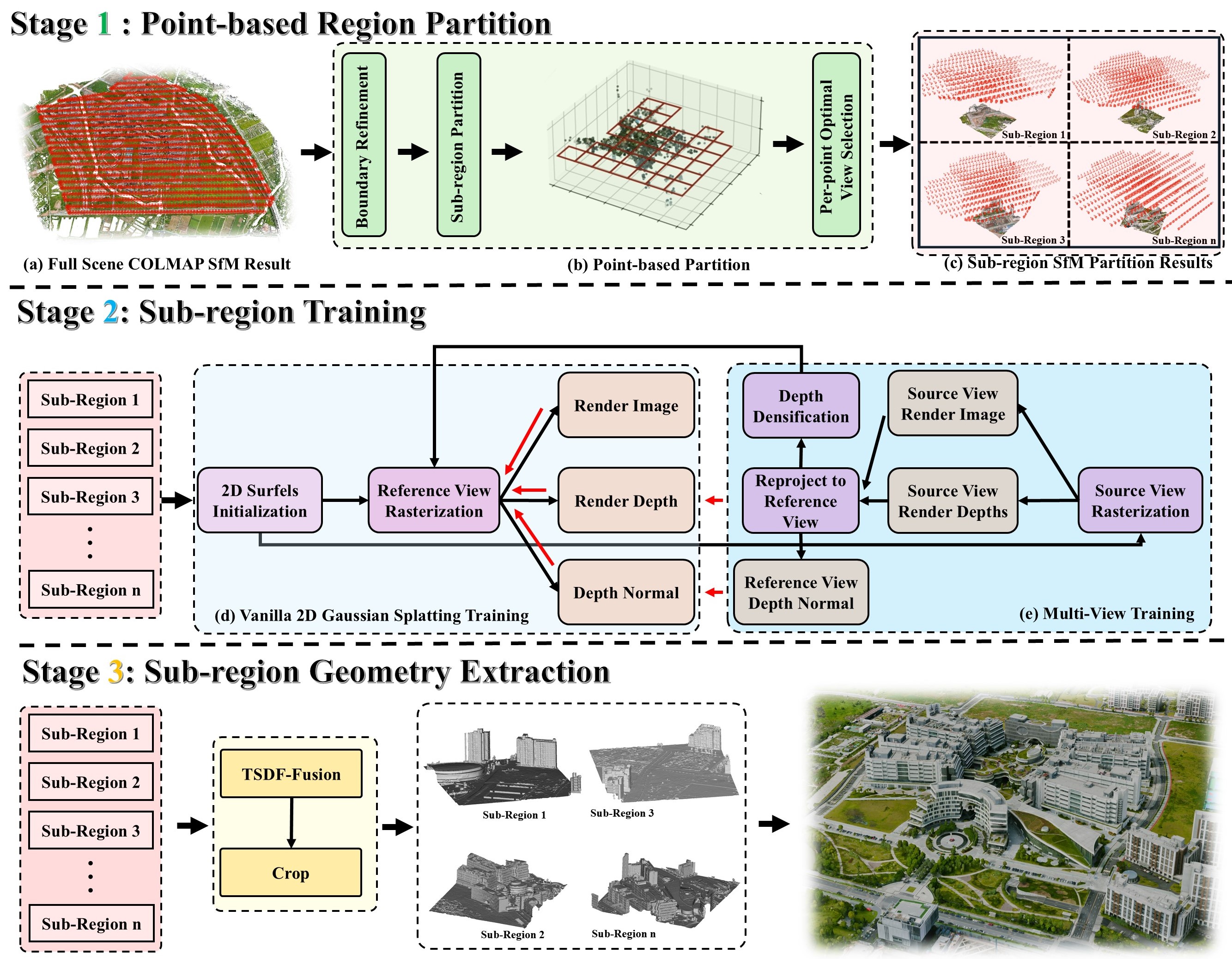}
    \caption{\textbf{ULSR-GS training pipeline.} Stage 1: Scene partitioning based on point cloud.  Stage 2: Training strategy based on multi-view geometric consistency. Stage 3: Extract the geometry of sub-regions separately and seamlessly merge. }
    \label{fig:teaser}
\end{figure*}

\section{Preliminaries} 
\subsection{Multi-View Pair Selection}
The\textit{ multi-view pair selection} utilizes a reference image along with three source images (\(N = 3\)) \cite{wu2024gomvs_mvs,yao2018mvsnet,yu2020fastmvs,su2023edgemvs_mvs, chen2019pointmvs}. For each image pair, they calculate a score \(s(i,j)\) based on sparse points. The score is defined as:
\begin{equation}
s(i,j) = \sum_p G(\theta_{ij}(p))
\label{score}
\end{equation}
where \(p\) represents a common track point in both views \(i\) and \(j\), and \(\theta_{ij}(p)\) is the baseline angle between the views.

The function \(G\) is a piecewise Gaussian function that favors a certain baseline angle \(\theta_0\):
\begin{equation}
    G(\theta) = 
\begin{cases} 
\exp\left(-\frac{(\theta - \theta_0)^2}{2\sigma_1^2}\right),  \theta \leq \theta_0 \\
\exp\left(-\frac{(\theta - \theta_0)^2}{2\sigma_2^2}\right), \theta > \theta_0 
\end{cases} 
\label{eq:G}
\end{equation}

\subsection{Depth Calculation in 2D Gaussian Splatting}
In 2DGS \cite{huang20242dgs}, two methodologies are employed for aggregating depth information.

\noindent\textbf{Weighted aggregation.} For each Gaussian $G_i$ intersecting along a viewing ray, the depth $D_{mean}$ is computed as a weighted sum:
\begin{equation}
    D_{mean}= \sum_i{\omega_iD_i}/(\sum_i{\omega_i}+\epsilon)
    \label{qe:meandepth}
\end{equation}
where $\omega_i$ represents the weight contribution of the $i$-th Gaussian. The result is normalized by the cumulative alpha values to obtain the expected depth.

\noindent\textbf{Maximum visibility.} This method focuses on selecting the depth of the most "visible" Gaussian along the viewing ray. The depth $D_{median}$ is determined by the first Gaussian that contributes significantly to the cumulative alpha: 

\begin{equation}
    D_{median}=\mathrm{max} \left \{ z_i|T_i>0.5 \right \} 
    \label{mediandepth}
\end{equation}
where $T_i$ quantifies the visibility based on the $i$-th Gaussian's alpha value. 

$D_{mean}$ facilitates smoother depth calculations, whereas $D_{median}$ more effectively captures rich geometric detail. However, it is highly susceptible to floating-point errors that can impact depth computation. In this paper, we adopt $D_{mean}$ as the depth representation for subsequent analyses. 

\section{Method}
Figure \ref{fig:teaser} illustrates the overall pipeline of our proposed ULSR-GS, which consists of three stages: \textbf{Stage 1: Multi-View Optimized Point Partitioning (Sec. \ref{pp}).} Starting from a full-scene COLMAP SfM result, we apply density-aware boundary refinement (Sec. \ref{boundary}) followed by spatial subdivision of the sparse point cloud into multiple rectangular sub-regions (Sec. \ref{view}). Each point is then assigned to its most informative image group via our proposed per-point optimal view selection, resulting in a well-structured sub-region configuration suitable for parallel processing (Sec. \ref{pview}). \textbf{Stage 2: Sub-region Training.}
Each sub-region is independently trained using 2DGS rasterization. Our training module includes an adaptive multi-view guided densification mechanism (Sec. \ref{MVD}), where source view depths are aggregated (Sec. \ref{dagg}) and reprojected to the reference view through an adapted window mask (Sec. \ref{add}). To enforce geometric consistency across depth and normal maps, we further introduce two multi-view regulations (Sec. \ref{loss}). This enables accurate local optimization with fine-grained geometry reconstruction. \textbf{Stage 3: Sub-region Geometry Extraction.}
After training, surface meshes for each sub-region are extracted using TSDF fusion followed by a cropping step. These independently reconstructed regions are then seamlessly assembled into a complete large-scale urban mesh without requiring additional global fusion.
\begin{figure*}[ht!]
    \centering
    \includegraphics[width=1\linewidth]{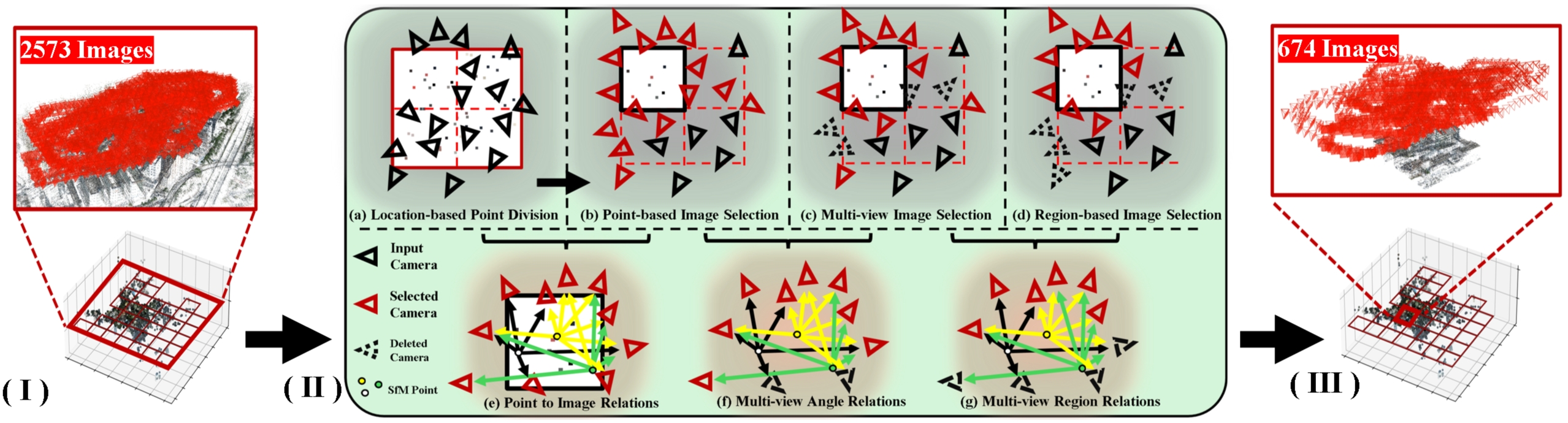}
    \caption{\textbf{Point-based partition strategy of ULSR-GS.} \textbf{(I)} We partition the training scene based on the positions of the point cloud projected onto the xz-plane. \textbf{(II)} For each sub-scene: \textit{(a)} we first select all matching images within the region; \textit{(b)} based on the camera relationships of the sub-scene, we select the corresponding images for each point; \textit{(c)} following the principle of multi-view optimal matching \cite{yao2018mvsnet}, we filter out images with poor view angles; and finally, \textit{(d)} considering the distance relationship, we further remove images that are too distant. \textbf{(III)} Examples of a sub-region significantly reduce the number of training images. }
    \label{fig:pipe}
\end{figure*}
\subsection{Multi-View Optimized Point Partitioning} \label{pp}
In this section, we will detail the point-based scene partitioning method in Figure \ref{fig:teaser} \textbf{Stage 1} and the method to select views for the partitioned sub-regions.
Different from previous studies that use drone photo locations for region partitioning \cite{vastgaussian,yuchen2024dogaussian,liu2024citygaussian}, our scene partitioning method is based on the initial point cloud of the scene and selects the best training images for the point cloud in each sub-region. The advantage of this method is that we can determine the boundaries of the mesh to be extracted for each sub-area at the early stage, without the need to merge the full scene. As illustrated in Figure \ref{fig:pipe}, we optimize the partition process by focusing on the most relevant parts of the scene.

\subsubsection{Density-controlled boundary refinement.}\label{boundary}
To effectively partition the initial points into $n \times n$ sub-regions, it is crucial to isolate the primary structural components of the scene and eliminate sparse and noisy SfM points that could distort the boundary definition. 

We first remove all 3D points with a SfM reprojection error \cite{schoenberger2016colmap} greater than a threshold $\epsilon_{error}$ ($\epsilon_{error} > 1.5$ in our experiments). This step cleans up the point cloud by discarding unreliable points that could skew the partitioning. Then we partition the 3D space into a voxel grid $v_i$ with size ($s=2$), and each SfM point $p_i = (x_i, y_i, z_i)$ is assigned to a voxel based on its coordinates:
\begin{equation}
    v_i = \left( \left\lfloor \frac{x_i}{s} \right\rfloor, \left\lfloor \frac{y_i}{s} \right\rfloor, \left\lfloor \frac{z_i}{s} \right\rfloor \right).
\end{equation}
For each voxel $v$, we count the number of points $N_v$ it contains:
\begin{equation}
    N_v = \sum_{i=1}^{N} \delta ( v_i, v ) ,
    \label{placeholder_label}
\end{equation}
where $N$ is the total number of points and $\delta(v_i, v)$ is the Kronecker delta function \cite{lang2020samplenet}. Then we compute a density threshold $N_\tau$ ($\tau = 33.3\%$) of the maximum voxel occupancy to identify the densely populated voxels: 
\begin{equation}
    N_\tau = \mathbf{max}_{v \in V} N_v.
\end{equation}
Voxels with occupancy $N_v > N_\tau$ are considered dense and are retained. Finally, we can get the precise boundary of the scene by computing the minimum and maximum coordinates of the points.

\subsubsection{Initial view selection.}\label{view}
As shown in Figure \ref{fig:pipe} (I), the input COLMAP SfM \cite{schoenberger2016colmap} point cloud is divided into $n \times n$ grids after density filtering. Each point in the sub-region is the feature of the images in which it was detected and matched. We first select all matched images as a coarse view selection (see Figure \ref{fig:pipe} (e)).

\subsubsection{Source view selection.} To further refine the view selection in Eq. \eqref{score}, a region constraint is applied based on the camera pair distance $d_{ij}$. Only image pairs with a distance below a specified maximum threshold $D_{\max}$ are considered for matching. 
The final matching score $S_{ij}$:
\begin{equation}
    S_{ij} = 
    \begin{cases} 
        \sum_p G(\theta_{ij}(p)) , \space \&  \space d_{ij} \leq D_{\max} \\ 
        0,  \text{otherwise}
    \end{cases}
    \label{eq:conditional_expression}
\end{equation}
Finally, for each reference image ${I}_{\text{ref}_i}$, the top $3$ source images {${I}_{\text{src}_i^1}$, ${I}_{\text{src}_i^2}$, ${I}_{\text{src}_i^3}$} with the highest matching scores $S_{ij}$ are selected to form the optimal set of views. 

\subsubsection{Per-point optimal view selection.} \label{pview}

Our goal is to ensure that each SfM point within a sub-region is associated with its most informative and geometrically robust image pairs. 
For each point $P_k \in \mathcal{P}$, we consider all possible groups of images that observe $P_k$. Specifically, each group $\mathcal{G}_i = { I_{\text{ref}_i}, I_{\text{src}_i^1}, I_{\text{src}_i^2}, I_{\text{src}_i^3} }$ consists of one reference image and three corresponding source images. Our method involves the basic \textit{W2C} steps.

We project $P_k$ onto the 2D image planes of each image $I_j \in \mathcal{G}_i$: 
\begin{equation}
\begin{bmatrix} u_{k_j} \\ v_{k_j} \\ 1 \end{bmatrix} =\frac{1}{Z_c} \begin{bmatrix}f_{x_j} & 0&c_{x_j}&0 \\ 0&f_{y_j}&c_{y_j}&0 \\ 0&0&1&0 
\end{bmatrix} \begin{bmatrix} R_j &t_j \\ 0^T&1 \end{bmatrix}\begin{bmatrix} X_{k_j} \\ Y_{k_j} \\ Z_{k_j}\\1 \end{bmatrix}\label{uv}
\end{equation}
where ($u_{k_j},v_{k_j}$) are the normalized pixel coordinates of $I_j$, $f_{x_j}$ and $f_{x_j}$ are the focal lengths of camera $I_j$ along the $x$ and $y$ axes, respectively. $c_{x_j}$ and $c_{y_j}$ are the image center coordinates. $R_j \in SO(3)$   and $t_j$ are the rotation matrix and translation vector of $I_j$ respectively. We then compute the Euclidean distance between the projected point ($u_{k_j},v_{k_j}$) and the principal point (image center) ($c_{x_j}$,$c_{y_j}$) of image $I_j$:
\begin{equation}
    d_{k_j}=\sqrt{(u_{k_j}-c_{x_j})^2+(v_{k_j}-c_{y_j})^2}.
\end{equation}
 
For each image group $\mathcal{G}_i$ observing $P_k$, we we calculate the average Euclidean distance:
\begin{equation}
    D_{avg}(P_k,\mathcal{G}_i) = \frac{1}{|\mathcal{G}_i|}\sum_{I_j \in \mathcal{G}_i} d_{k_j}
\end{equation}
Among all possible four-image groups that observe $P_k$, we select the group with the smallest $D_{avg}(P_k)_{min}$. This ensures that $P_k$ is primarily reconstructed using images where it is closest to the image centers in their respective 2D planes, thereby enhancing the reliability of its triangulation.
Since we want to select the optimal image group for every initial SfM point in the sub-regions. We simply identify the group for $P_k$ $\mathcal{G}$ by minimizing  $D_{avg}$:
\begin{equation}
    \mathcal{G}_{opt}(P_k)= agr\space \min_{\mathcal{G}_i} D_{avg}(P_k,\mathcal{G}_i).
\end{equation}
This ensures that $P_k$ is reconstructed using images where it appears closest to the image centers, enhancing reconstruction reliability due to reduced distortion and higher image quality near the center.  After determining $\mathcal{G}_{\text{opt}}(P_k)$for all points $P_k \in \mathcal{P}$, we aggregate the utilized images $I_{used}$ and exclude any images not present: 
\begin{equation}
    I_{used} = \bigcup_{P_k \in \mathcal{P}} \mathcal{G}_{opt}(P_k)
\end{equation}

After determining the best image groups for all 3D points within the sub-region, we exclude any images that do not appear in any of the selected best groups. In our experiments, the excluded images are mainly located at the outermost sides of the sub-region, which means they are redundant images which can only observe a few points in the region.

\subsubsection{Scene Partitioning Strategy}
To address the issue where some sub-regions may contain too few points (especially in boundary areas) after dividing the scene into smaller parts, we evaluate each sub-region as follows: We retain a sub-region only if both the number of points and the number of matched images within it reach at least 10 \% of the average per-subregion counts  (i.e., the total number of points or images divided by $n^2$). 
\noindent\begin{table}[h]
    \centering
   \small
   \caption{\textbf{Example Per Scene Partitioning Strategy.} }
    \begin{tabular}{l|c|c|c}
        \toprule
Scene & Images & Grid Size & Used \\
       \midrule
 Campus \cite{UrbanScene3D} & 5871 & $8\times8$ & 41 \\
 
 Residence \cite{UrbanScene3D} & 2582 & $6\times6$ & 20\\
  
 SZTU \cite{xiong2024gauuscene}& 1500 & $6\times6$ & 17 \\
  
 LOWER\_CAMP \cite{xiong2024gauuscene}& 670 & $4\times4$ & 12\\
        \bottomrule
    \end{tabular}
    
    \label{table:perscene}
\end{table}

We partition each scene based on the number of images to ensure efficient processing and optimal reconstruction quality. Specifically, we divide the regions according to the following criteria: 
\begin{enumerate}
    \item Scenes with fewer than 1,000 images are divided into a $4\times4$ grid.
    \item Scenes with fewer than 3,000 images are divided into a $6\times6$ grid.
    \item Scenes with more than 3,000 images are divided into an $8 \times 8$ grid.
\end{enumerate}

Table \ref{table:perscene} lists the examples of the number of initially partitioned sub-regions for each scene and the number of sub-regions that were ultimately included in the training after applying our filtering criteria based on point cloud density and image coverage.

\subsection{Adaptive Multi-View Densification} \label{MVD}
In this section, we detail how multi-view geometric consistency is exploited to introduce densification based on rendered depth maps for 2DGS. As demonstrated in Figure \ref{fig:teaser} (Stage 2), each subregion can be assigned to an independent GPU for training. However, vanilla 2DGS often struggles to capture fine geometry details due to densification based on single-view image gradients, resulting in low-fidelity rendering and overly smooth geometry extraction.  Therefore, it is necessary to add more fine Gaussian primitives for capturing high-frequency geometric details \cite{fan2024trim, li2024mvgsplatting, chen2024mvsplat, Yu2024GOF}. In ULSR-GS, we perform additional densification via an MVS-like approach to address the limitations of overly smooth meshes resulting from $D_{mean}$ computations in Eq. \ref{qe:meandepth} from TSDF fusion. This method projects $D_{mean}$ into 3D space \cite{yao2018mvsnet,cheng2024gaussianpro} and combines it with the RGB information of the GT image to enrich the Gaussian primitives.

\subsubsection{Multi-view depth aggregation.} \label{dagg} In our approach, we perform weighted averaging of depth from multiple source views. This weighting assigns a confidence score to each depth estimate based on its geometric consistency \cite{yao2018mvsnet}, ensuring that source views with higher consistency have a greater influence on the final depth estimation.

Since we have $3$ source views from Eq. \eqref{eq:conditional_expression}, with each rendered depth $D_{mean}$ denoted as $\mathbf{D}_{\text{src}_i}$ for $i = 1, 2, 3$, and the depth map of the reference view denoted as $\mathbf{D}_{\text{ref}}$. For each pixel $p_{\text{ref}}$ in the reference view, the final fused depth estimate $\mathbf{D}_{\text{final}}(p_{\text{ref}})$ is obtained through weighted fusion of the source views:
\begin{equation} 
\mathbf{D}_{\text{final}}(p_{\text{ref}}) = \frac{\sum_{i=1}^{N} w_{\text{src}_i}(p_{\text{src}_i}) \cdot \mathbf{D}_{\text{src}_i}(p_{\text{src}_i})}{\sum_{i=1}^{N} w_{\text{src}_i}(p_{\text{src}_i})}, 
\label{eq:depthf} 
\end{equation} 
where $\mathbf{D}_{\text{src}_i}(p_{\text{src}_i})$ is the depth value at the projected pixel $p_{\text{src}_i}$ in the $i$-th source view corresponding to $p_{\text{ref}}$. The weight $w_{\text{src}_i}(p_{\text{src}_i})$ is based on the geometric consistency score, measuring the reliability of the depth estimate from the $i$-th source view:
\begin{equation} w_{\text{src}i}(p_{\text{src}i}) = \exp \left( -\frac{E_{\text{depth}}(p_{\text{ref}}, p_{\text{src}i})}{\sigma^2} \right), \label{depth_weight
} \end{equation} where $E_{\text{depth}}(\cdot)$ represents the depth error between the reference view and the source view:
\begin{equation} 
E_{\text{depth}}(p_{\text{ref}}, p_{\text{src}_i}) = \left| \mathbf{D}_{\text{ref}}(p_{\text{ref}}) - \mathbf{D}_{\text{src}_i}(p_{\text{src}_i}) \right| \label{eq:depth_error} 
\end{equation}

\begin{table*}[ht]
    \footnotesize
    \setlength\tabcolsep{2pt}
    \centering 
    \renewcommand\arraystretch{1.2}
    \caption{\textbf{Quantitative results on GauU-Scene dataset \cite{xiong2024gauuscene}.} We report precision, recall, and F-1 score at the threshold $\tau = 0.025$ (relative). The \textbf{best} and \underline{second}  results are highlighted. "OOM" denotes no results due to GPU out of memory.  " *" denotes no result due to no convergence. } 
    \resizebox{\textwidth}{!}{
    \begin{tabular}{c|ccc|ccc|ccc|ccc|ccc|ccc|ccc|c} % 16列
        \toprule
        Scene & \multicolumn{3}{c|}{LOWER\_CAMP} & \multicolumn{3}{c|}{UPPER\_CAMP}  &\multicolumn{3}{c|}{LFLS} &\multicolumn{3}{c|}{SMBU}  & \multicolumn{3}{c|}{SZIIT}& \multicolumn{3}{c|}{SZTU}&  \\
        \midrule
        Metrics ($\tau$ = 0.025)& Pre.$\uparrow$ & Rec.$\uparrow$ & F1$\uparrow$ & Pre.$\uparrow$ & Rec.$\uparrow$ & F1$\uparrow$& Pre.$\uparrow$ & Rec.$\uparrow$ & F1$\uparrow$& Pre.$\uparrow$ & Rec.$\uparrow$ & F1$\uparrow$& Pre.$\uparrow$ & Rec.$\uparrow$ & F1$\uparrow$ & Pre.$\uparrow$ & Rec.$\uparrow$ & F1$\uparrow$ &Time \\
        \midrule
        Neuralangelo  &
        * &* &* & 
        * &* &* & 
        * &* &* & 
        * &* &* & 
        * &* &* & 
        * &* &* & 
        $>$100 h\\ 
        \midrule
        SuGaR & 
        0.279 & 0.263 & 0.271 & 
        0.261 & 0.273 & 0.267 & 
        0.271 & 0.300 & 0.285 & 
        0.376 & 0.409 & 0.392 & 
        0.249 & 0.276 & 0.263 & 
        0.272 & 0.299 & 0.285 &   4.1h \\
        2DGS-60K & 
        0.583 & \textbf {0.693} & 0.633 & 
        0.554 & 0.563 & 0.559 & 
         0.543 &  0.557 &  0.550 & 
        0.580 & 0.745 & 0.652 & 
        0.584 & 0.617 & 0.600 & 
        \underline{0.596} &\underline {0.613} & \underline{0.604} &  \textbf{2.1h} \\
        
        GOF-60K & 
         0.667 &  0.652 &  0.659 & 
        \textbf{0.693} &  0.672 & \underline {0.682} & 
        \underline{0.607} & \underline{0.659} & \underline{0.631} & 
        \underline{0.679} &   0.784 & \underline{0.727} & 
         0.649 & \underline{0.695} & \underline{0.671} & 
        OOM & OOM & OOM  &  5.3h\\
        PGSR-60K & 
        \textbf{0.683} & 0.648 &\underline {0.665} & 
         0.672 & \underline{0.673} &  0.672 & 
        OOM & OOM & OOM & 
         0.660 & \textbf{0.795} &  0.721 & 
        \underline {0.679} &  0.642 &  0.659 & 
         0.548 &  0.542 &  0.545 &   5.7h \\
        \midrule
        Ours  & 
        \underline{0.681} & \underline{0.689} & \textbf{0.685} & 
        
        \underline{0.690} & \textbf{0.705} & \textbf{0.697} & 
        
        \textbf{0.684} & \textbf{0.703} & \textbf{0.693} & 
        
        \textbf{0.737} & \underline{0.789} &  \textbf{0.762} & 
        
        \textbf{0.705} & \textbf{0.713} & \textbf{0.70}9 & 
        
        \textbf{0.713} & \textbf{0.723} & \textbf{0.718} & \underline{3.7h}\\

        \bottomrule
    \end{tabular} }
    
    \label{tab:mesh}  
\end{table*}

\begin{table*}[ht]
   \footnotesize
    \setlength\tabcolsep{1pt}
    \renewcommand\arraystretch{1.2}
    \centering 
    \caption{\textbf{Quantitative results on Matrix City dataset (Small City Scene) \cite{li2023matrixcity} and two custom-collected datasets (Scene1 and Scene2).} We report precision, recall, and F-1 score at different distance thresholds $\tau = 1.0m$ and $\tau = 0.5m$ (metric). The \textbf{best}, \underline{second} results are highlighted. "OOM" denotes no results due to GPU out of memory. Note that "City-GS V2+Ours(2$\times$2)" means we train CityGaussianV2 \cite{liu2024citygaussianv2efficientgeometricallyaccurate} on our proposed partition method. } 
    \resizebox{\textwidth}{!}{
    \begin{tabular}{l|ccc|ccc|ccc|ccc|ccc|ccc} % 16列
        \toprule
         & \multicolumn{6}{c|}{Matrix City} & \multicolumn{6}{c|}{Scene1} & \multicolumn{6}{c}{Scene2} \\
        \midrule
        & \multicolumn{3}{c|}{$\tau=1.0 m $} & \multicolumn{3}{c|}{$\tau=0.5 m$} & \multicolumn{3}{c|}{$\tau=1.0 m $} & \multicolumn{3}{c|}{$\tau=0.5 m$}& \multicolumn{3}{c|}{$\tau=1.0 m $} & \multicolumn{3}{c}{$\tau=0.5 m$}
         \\ 
        \midrule
        Metrics & Pre.$\uparrow$ & Rec.$\uparrow$ & F1$\uparrow$& Pre.$\uparrow$ & Rec.$\uparrow$ & F1$\uparrow$& Pre.$\uparrow$ & Rec.$\uparrow$ & F1$\uparrow$& Pre.$\uparrow$ & Rec.$\uparrow$ & F1$\uparrow$& Pre.$\uparrow$ & Rec.$\uparrow$ & F1$\uparrow$& Pre.$\uparrow$ & Rec.$\uparrow$ & F1$\uparrow$ \\
        \midrule
           
        2DGS-60K (Baseline)&0.551 &0.692 &0.613 &0.331 &0.502 &0.398& OOM & OOM & OOM & OOM & OOM & OOM & OOM & OOM & OOM & OOM & OOM & OOM\\
        City-GS V2 &\textbf{0.871} & \underline{0.901} & \underline {0.886} &\textbf{0.667} &\underline{0.753} &\textbf{0.707} & OOM & OOM & OOM & OOM & OOM & OOM & OOM & OOM & OOM & OOM & OOM & OOM\\
        City-GS V2+Ours(2$\times$2) &- &- &- &- &- &- & \textbf {0.845} & \underline {0.875} &\underline {0.860} &\textbf {0.646} &\underline {0.753} &\underline {0.695} & \underline{0.702} & \underline{0.746} &\underline {0.723} & \textbf{0.569} &\underline{0.637} & \underline{0.601}\\
        \midrule
        Ours & \underline {0.869} & \textbf {0.916} & \textbf {0.892} & \underline {0.660} & \textbf {0.759} & \underline {0.706} & \underline {0.837}& \textbf {0.929} & \textbf {0.880} & \underline     {0.632} & \textbf {0.787} & \textbf {0.701} & \textbf{0.710} & \textbf{0.748} & \textbf{0.728} & \textbf{0.569} & \textbf{0.651} & \textbf{0.607}\\
        \bottomrule
    \end{tabular} }
    
    \label{tab:mesh1}  
\end{table*}

\subsubsection{Adaptive depth densification.} \label{add} Directly projecting all depth as new 2D Gaussian primitives after checking the geometric consistency introduces excessive redundant information into the training scene \cite{li2024mvgsplatting}. This redundancy negatively impacts both the training speed and the accuracy of the scene representation. To address this issue, we introduce an adaptive densification window mask that confines the densification area. This approach eliminates inaccurate values at the edges of the depth map and adaptively accounts for non-uniform depth changes caused by varying viewpoint poses.

To determine the mask size adaptively, we base it on the depth gradient $|\nabla \mathbf{D}(x)|$, which represents the rate of change in depth at each pixel $x$. The window size is inversely proportional to the mean gradient $\bar{g}$, implying that regions with higher depth variations have smaller windows to capture finer details, while regions with lower variations have larger windows. We first compute the mean gradient $\bar{g}$ over the entire depth map of size $h \times w$:
\begin{equation} \bar{g} = \frac{1}{h \times w} \sum_{x \in \mathbb{D}} |\nabla \mathbf{D}(x)| \label{eq:gradient
} \end{equation}
Based on the average gradient $\bar{g}$, we define the height and width of the window to dynamically adapt to the depth changes in the scene:
\begin{equation} (h_{\text{win}}, w_{\text{win}}) = \frac{k}{\bar{g} + \epsilon} \cdot \frac{(h, w)}{2} \label{eq:window
} \end{equation}
where $k$ and $\epsilon$ are proportional constants that control the change in window size. The term $\epsilon$ prevents division by zero when the gradient is very small.

We then project the depth $\mathbf{D}_{\text{final}}$ within the window $W(u, v) \in (h_{\text{win}}, w_{\text{win}})$ after the geometric consistency check and fusion from Eq. \eqref{eq:depthf}:
\begin{equation} \mathbf{P}_w(u, v) = \mathbf{D}_w(u, v) \cdot \mathbf{K}_p^{-1} \begin{bmatrix} u \ v \ 1 \end{bmatrix} \label{eq
} \end{equation}
where $\mathbf{K}_p^{-1}$ is the inverse of the intrinsic camera matrix, and $\mathbf{P}_w(u, v)$ represents the 3D point projected from the windowed depth map.
Following the settings of MVG-Splatting \cite{li2024mvgsplatting}, we perform additional rescaling and rotational alignment on the newly added Gaussian primitives after each densification step to ensure consistency across the scene.

\subsection{Loss Functions} \label{loss}

\noindent\textbf{Geometric consistency loss.} We optimize the geometric consistency between the reference view and the source views via the \textit{reprojection error} that compares the reprojected reference depth $\mathbf{D}_{\text{ref}}$ with the source depth $\mathbf{D}_{\text{src}}$ in the source view as defined in Eq. \eqref{eq:depth_error}.

\noindent\textbf{Multi-view normal consistency loss.} For each projected point, we compute the angular error between the normal vectors of the reference view $\mathbf{N}_{\text{ref}}$ and the source view $\mathbf{N}_{\text{src}}$. This normal vector consistency ensures that the geometry of the primitive  remains consistent under different viewing angles:
\begin{equation} E_{\text{normal}} = 1 - \frac{\mathbf{N}_{\text{ref}} \cdot \mathbf{N}_{\text{src}}}{|\mathbf{N}_{\text{ref}}| |\mathbf{N}_{\text{src}}|} \label{eq1
} \end{equation}

\noindent\textbf{Final loss function.} The final loss function is defined as:
\begin{equation} L = \alpha \cdot E_{\text{depth}} + \beta  \cdot E_{\text{normal}} + L_{\text{geo}} + L_r \end{equation}
where $\alpha$, $\beta$, are weighting factors that balance the contributions of each error term.  $L_{\text{2DGS}}$ includes two regularization components from 2DGS \cite{huang20242dgs}: depth distortion $\ell_d$ and normal consistency $\ell_n$.  $L_{\text{3DGS}}$ represents the RGB reconstruction loss, combining $L_1$ loss and the D-SSIM metric as used in 3DGS \cite{3dgs}.

\section{Experiments} \label{exp_setting}
\subsection{Implementation}
In our experiments, we initially employ COLMAP \cite{schoenberger2016colmap} to obtain the SfM results for the entire scene. Subsequently, we align the sparse point cloud using the Manhattan World Alignment, ensuring that the \textit{y-axis} is perpendicular to the \textit{xz} plane. We set the angular threshold $\theta_{min}$ = 90, and the distance threshold $D_{max}$ = $\sqrt{d_x\times d_y} $, where $d_x$ and $d_y$ are the distances of each sub-region.  After partitioning, we double the boundary of the initial point cloud for each region for the initialization. This configuration ensures that the edges of each subregion receive sufficient training information, thereby preventing edge mismatches during the stitching process. Redundant points are automatically pruned during training through opacity culling. Experiments for our method are conducted on 4 RTX 4090 GPUs, and each sub-region is trained individually by one GPU.

For each subregion, training is conducted for 40k iterations.  Adaptive multi-view densification begins at the 20k-th iteration and terminates at the 30k-th iteration. Regarding the loss function, we set the parameters $\alpha$ = 0.01 and $\beta$ = 0.1. All other loss functions and training parameters are configured identically to those used in 3DGS and 2DGS frameworks. After training, we employ TSDF \cite{werner2014tsdf} to extract the mesh of each sub-region then crop using their partition bounding boxes and subsequently stitched together.

\begin{figure}[!h]
    \centering
    \includegraphics[width=1\linewidth]{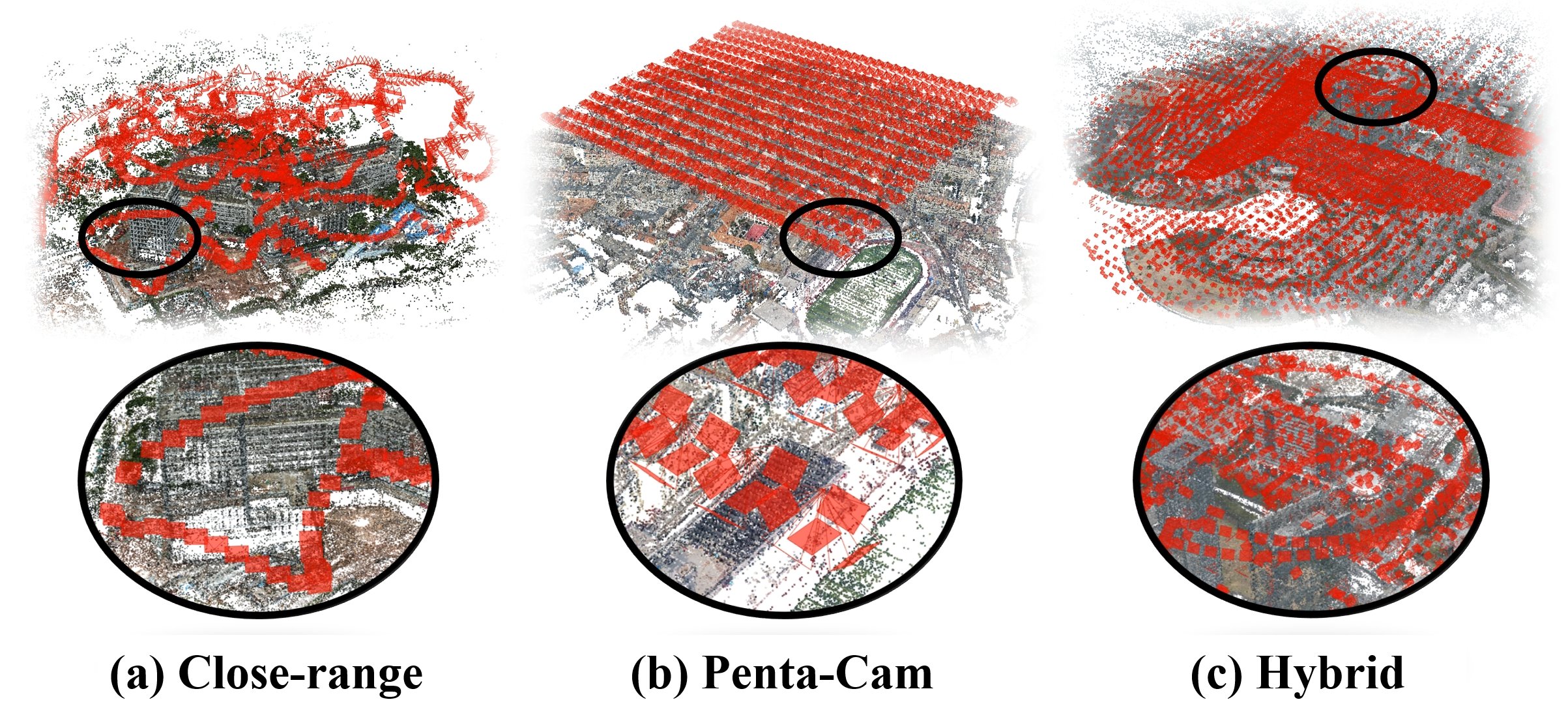}
    \caption{\textbf{Examples of experiment datasets.} Our experiments are conducted on different types of aerial oblique photography datasets, such as (a) close-range photogrammetry, (b) five-directional (penta-cam) photogrammetry, and (c) hybrid photogrammetry which contains both close-range and penta-cam photogrammetry.}
    \label{fig:oblique}
\end{figure}

\begin{figure*}[!hb]
    \centering
    \includegraphics[width=1\linewidth]{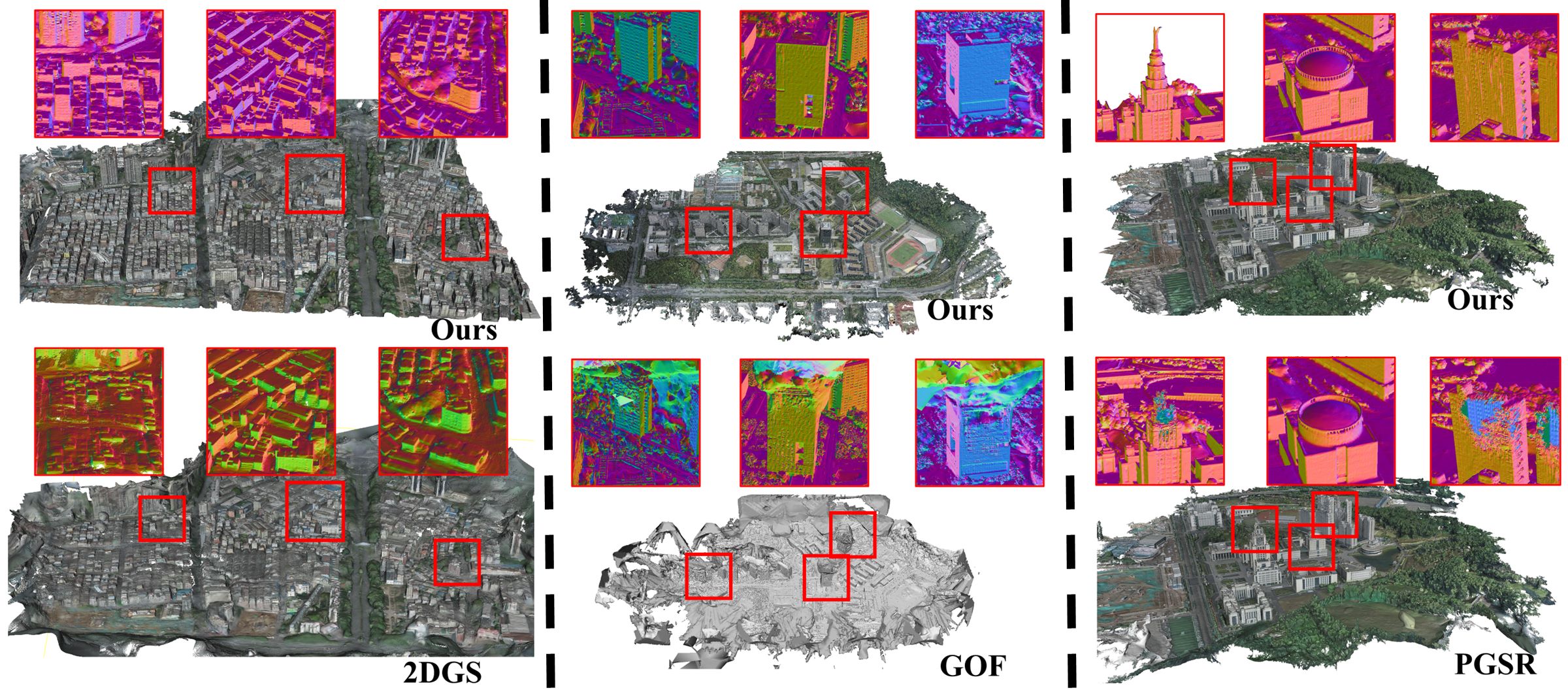}
    \caption{\textbf{Qualitative comparison with single GPU GS-based methods.} We show the full scene comparison with 2DGS \cite{huang20242dgs}, GOF\cite{Yu2024GOF}, and PGSR\cite{chen2024pgsr}. From left to right are the \textit{LFLS}, \textit{SZIIT}, and\textit{ SMBU} scenes of the GauU-Scene \cite{xiong2024gauuscene} dataset.}
    \label{fig:full}
\end{figure*} 
\begin{figure*}[ht]
    \centering
    \includegraphics[width=1\linewidth]{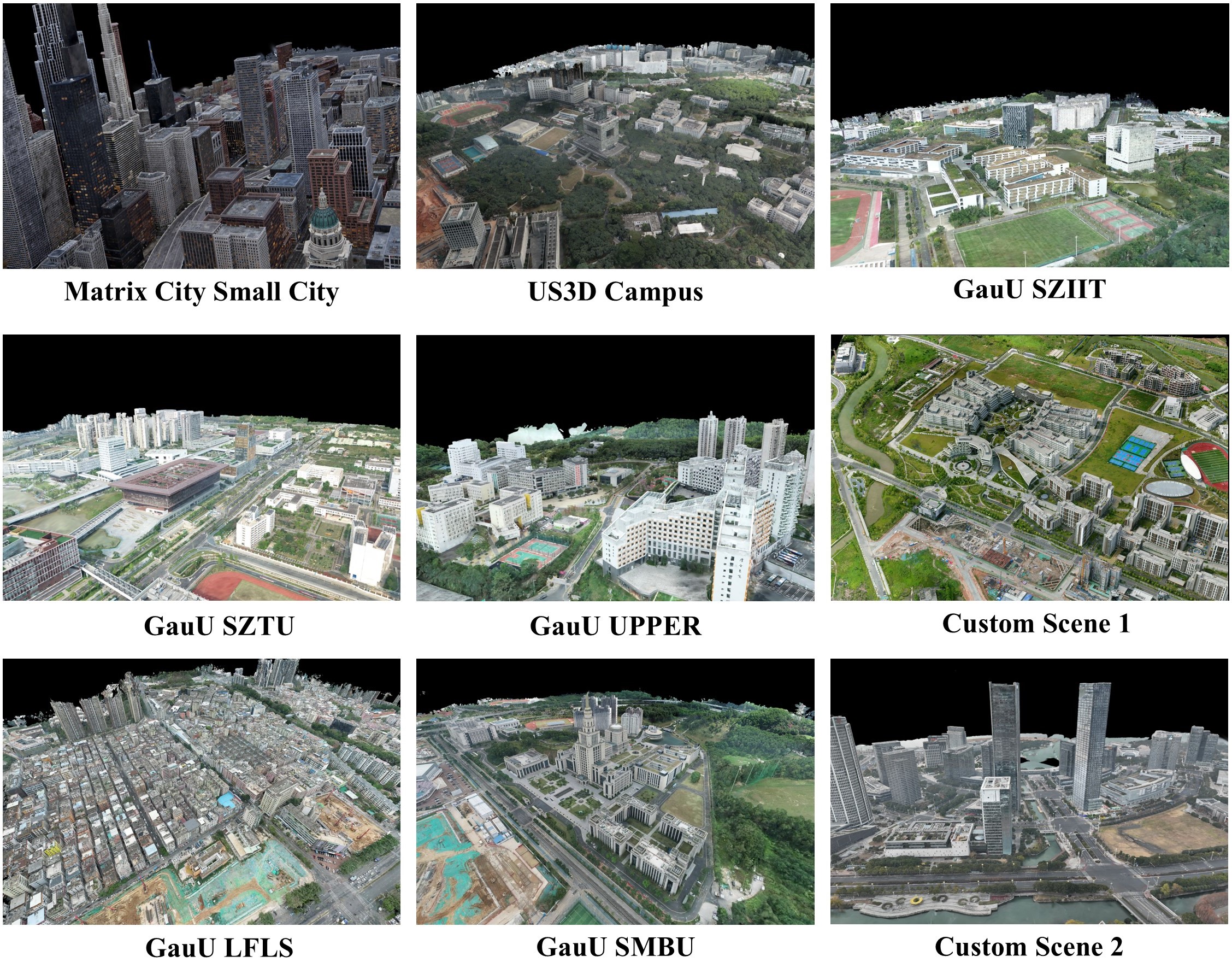}
    \caption{\textbf{Rendering Rensults.} We demonstrate the full-scene (zoomed-out) rendering detail of the selected scenes.}
    \label{fig:render}
\end{figure*}
\subsection{Datasets}
As shown in Figure \ref{fig:oblique}, our experiments are conducted on large-scale aerial oblique photogrammetry datasets: GauU-Scene (GauU) \cite{xiong2024gauuscene}, UrbanScene3D (US3D) \cite{UrbanScene3D}, and two custom-collected datasets, namely Scene 1 and Scene 2 (example rendering results can be found at Figure \ref{fig:render}). Both Scene 1 (Penta-Cam) and Scene 2 (Close-range + Penta-Cam) contain around 10,000 images. 

\begin{enumerate}
    \item GauU-Scene (GauU) \cite{xiong2024gauuscene} dataset is a large-scale dataset specifically for Gaussian Splatting. It contains six areas with an average coverage of more than 1 $km^2$. However, each scene only contains an average of 1,200 high-resolution drone oblique photography images with an average overlap rate of no more than 50$\%$. The lack of densification caused by sparse data will cause serious under-reconstruction, which poses a great challenge to the single-GUP-based GS modeling. We use the UAV-scanned LiDAR point cloud provided by GauU as the ground truth (GT) for the evaluation mesh. 

    \item UrbanScene3D (US3D) \cite{UrbanScene3D}  dataset is a large-scale urban dataset that is strictly collected according to the requirements of oblique photography. The 'Campus' scene contains 5,000 Penta-Cam images within 2.5 km, with an overlap rate of more than 70$\%$, which is a great challenge for the GS method with a single GPU, and also for the training strategy of multiple GPUs. It also contains two close-range oblique photogrammetry scenes, which poses a challenge to the partitioning strategy of the GS method with multiple GPUs.

    \item We also include two additional custom-collected datasets in the real world. Scene 1 has a measurement range of about 4$km^2$, and uses Penta-Cam to collect more than 8,000 high-resolution complete campus images. Scene 2 is a 2$km^2$ complex urban oblique photography dataset containing a large amount of water environment and skyscrapers. It uses Penta-Cam for overall collection, combined with close-range shooting to take separate supplementary photos of each skyscraper over 200 m. This uneven collection of 10,000 images is very close to the actual photogrammetry collection process, which poses a great challenge to all GS methods. We also use USV orthophoto LiDAR scanning to collect point clouds within the coverage area as ground truth.
\end{enumerate}

\subsection{Metrics}

For evaluating NVS rendering results, we employ PSNR, SSIM, and LPIPS on Urbanscene3D and GauU datasets. In assessing the reconstructed meshes, we report Precision, Recall, and F-score against the GT Lidar pointclouds on GauU-scene dataset, and two custom collected Scene1, and Scene2.  Following the settings of Vast Gaussian \cite{vastgaussian}, we adjust the training configurations for all GS-based methods for a fair comparison. The input training photos are downsampled by a factor of 4 (except for the Matrix City dataset \cite{li2023matrixcity}).

\subsubsection{Rendering quality evaluation}

\begin{itemize}
\item PSNR (Peak Signal-to-Noise Ratio): Measures the pixel-wise difference between the rendered images and the ground truth images:
\begin{equation}
\text{PSNR} = 10 \cdot \log_{10}\left(\frac{\text{MAX}_I^2}{\text{MSE}}\right)
\end{equation}
where  is the maximum possible pixel value and MSE is the Mean Squared Error.

\item SSIM (Structural Similarity Index Measure): Evaluates perceived image quality based on luminance, contrast, and structure:
\begin{equation}
    \text{SSIM}(x,y) = \frac{(2\mu_x\mu_y + C_1)(2\sigma_{xy} + C_2)}{(\mu_x^2 + \mu_y^2 + C_1)(\sigma_x^2 + \sigma_y^2 + C_2)}
\end{equation}

\item LPIPS (Learned Perceptual Image Patch Similarity): Measures perceptual similarity by comparing features extracted from pretrained neural networks:
\begin{equation}
    \text{LPIPS}(x,y) = \sum_l \frac{1}{H_lW_l} \sum_{h,w} \Vert w_l \odot (\hat{y}_l^{hw}-\hat{x}_l^{hw})\Vert_2^2
\end{equation}

\end{itemize}
\subsubsection{Mesh quality evaluation}
Since GauU-scene \cite{xiong2024gauuscene} does not provide a standard evaluation code, we follow the Tanks and Tample \cite{knapitsch2017tanks} evaluation pipeline to quantitatively assess reconstructed meshes against GT LiDAR point clouds:

\begin{enumerate}
    \item To focus the evaluation on relevant regions, we compute the overlapping axis-aligned bounding box of both point clouds and mesh and crop them accordingly. We then perform Iterative Closest Point \cite{icp} (ICP) alignment to minimize misalignment between the reconstructed and GT point clouds.
    \item We uniformly sample number of points from the reconstructed mesh and GT points to 50 million.
    \item We utilize \textit{KDTreeFlann} to compute precision, recall, and F-score using predefined distance thresholds.

    \item Precision (Pre): Evaluates the accuracy of reconstructed surfaces against ground truth:
    \begin{equation}
    \text{Precision} = \frac{|{p \in \mathcal{P}{\text{rec}} | d(p, \mathcal{P}{gt}) < \tau}|}{|\mathcal{P}_{\text{rec}}|}
    \end{equation}

    \item Recall (Rec): Measures the completeness of reconstruction:
    \begin{equation}
    \text{Recall} = \frac{|\{p \in \mathcal{P}_{gt} | d(p, \mathcal{P}_{\text{rec}}) < \tau\}|}{|\mathcal{P}_{gt}|}
    \end{equation}

    \item F-1 Score: Provides a balanced measure combining Precision and Recall:
    \begin{equation}
        \text{F1 Score} = 2 \times \frac{\text{Precision} \times \text{Recall}}{\text{Precision} + \text{Recall}}
    \end{equation}
\end{enumerate}

\subsection{Compared methods.} 
We compare our proposed ULSR-GS with: (1) single-GPU-based GS methods capable of mesh extraction, including SuGaR \cite{guedon2023sugar}, 2DGS \cite{huang20242dgs}, GOF \cite{Yu2024GOF}, and PGSR \cite{chen2024pgsr}; (2) multi-GPU-based GS method capable of mesh extraction CityGaussianV2 \cite{liu2024citygaussianv2efficientgeometricallyaccurate}; (3) open source or free-to-use MVS-based photogrammetry projects, including COLMAP and Reality Capture; and (4) we also compare Vast Gaussian's \cite{vastgaussian} partiton strategy on rendering tasks and grometry task.

% For NVS rendering evaluation, we include NeRF-based methods Mega-NeRF \cite{turki2022mega} and Switch-NeRF \cite{mi2023switchnerf}, as well as the GS-based 3DGS \cite{3dgs}, Mip-Splatting \cite{yu2024mipsplatting} and Vast Gaussian \cite{vastgaussian}.

\begin{figure*}[!hbp]
    \centering
    \includegraphics[width=1\linewidth]{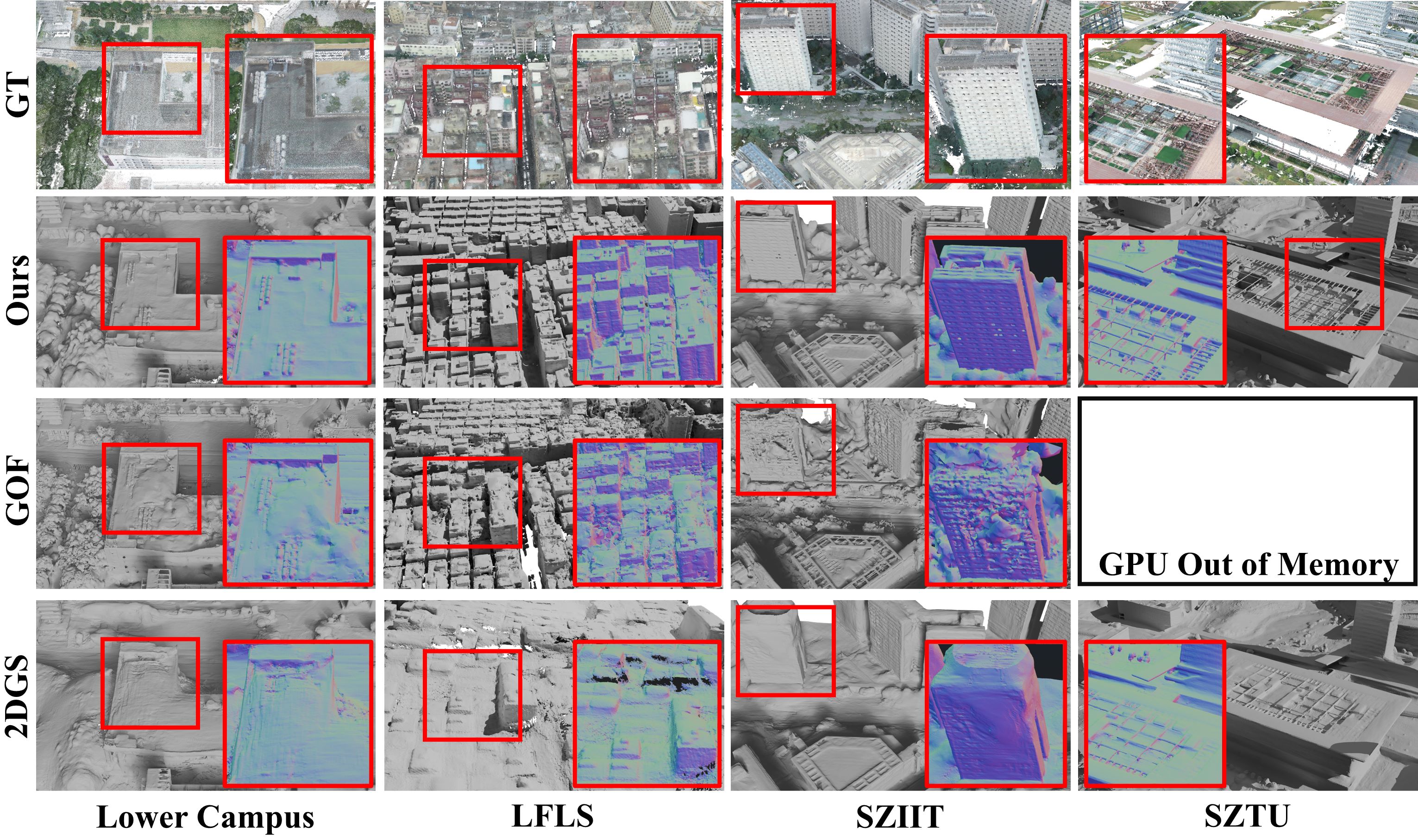}
    \caption{\textbf{Qualitative comparison on the GauU-Scene dataset \cite{xiong2024gauuscene}. } We demonstrate the zoomed-in surface normal detail of the selected areas in red box.}
    \label{fig:mesh_comp}
\end{figure*}

\begin{figure}
    \centering
    \includegraphics[width=1\linewidth]{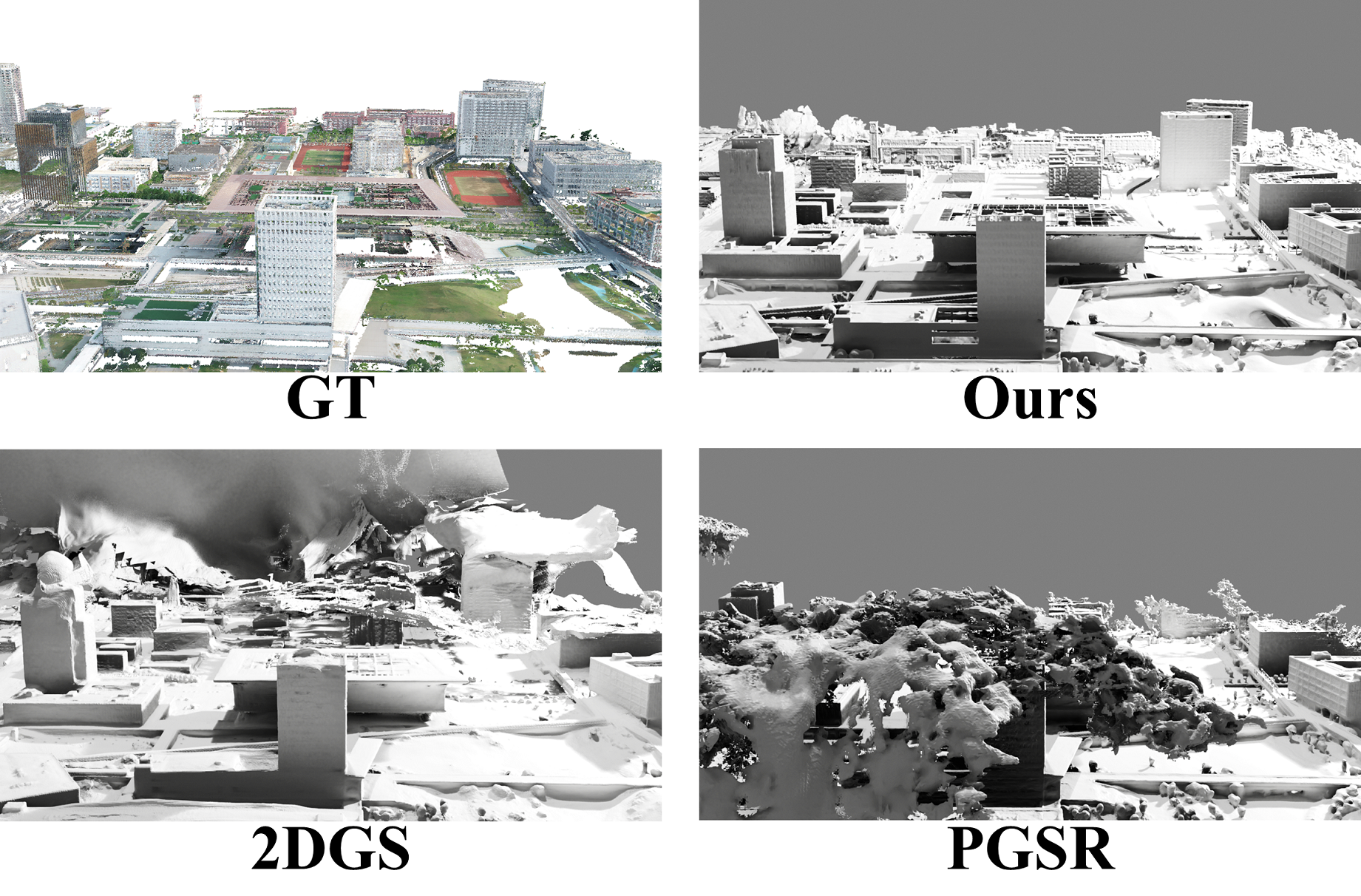}
    \caption{\textbf{Qualitative comparison on the GauU-Scene \cite{UrbanScene3D}dataset \textit{SZTU} scene.} When the scene range expands and the number of images increases, the geometric reconstruction quality of 2DGS \cite{huang20242dgs} and PGSR \cite{chen2024pgsr} drops sharply.}
    \label{fig:sztu}
\end{figure}

\subsection{Experiment results.} \label{results}

\subsubsection{Comparison with single-GPU-based GS Methods.}

The comparison with single-GPU methods is conducted on the GauU-Scene \cite{xiong2024gauuscene} dataset. It is still important to note that the comparison with these methods is only as fair as possible. We increase the training iterations  of all single-GPU methods by a factor of 2 as much on hardware with 24GB VRAM (a factor of 3 or more will cause most methods to GPU out-of-memory (OOM)), and all training parameters were adjusted accordingly. The full-scene results clearly indicate that our method achieves reconstructions with substantially greater detail and completeness than these baselines. 

Figure \ref{fig:full} provides a qualitative comparison of ULSR-GS against three recent single-GPU reconstruction methods: 2DGS \cite{huang20242dgs}, GOF \cite{Yu2024GOF}, and PGSR \cite{chen2024pgsr}. In particular, the 2DGS \cite{huang20242dgs} approach produces excessively smooth mesh surfaces that fail to capture fine-grained architectural details, especially in densely built areas. As a result, it cannot accurately reconstruct the intricate geometry of building structures in such regions. The GOF \cite{Yu2024GOF} method, which employs a marching tetrahedron algorithm for mesh extraction, often yields meshes with spurious connections to the sky. These artifacts disrupt the continuity of the model and prevent GOF from producing a correct, complete outline of the urban scene. Although PGSR \cite{chen2024pgsr} leverages multi-view geometric constraints similar to our method, it selects source views based on those closest in viewpoint to the reference image. In aerial oblique photography scenarios, this strategy frequently chooses source images whose content diverges significantly from the reference view’s observations, hindering effective multi-view geometric consistency enforcement. Consequently, as shown in Figure \ref{fig:sztu} PGSR’s reconstructions suffer from reduced mesh integrity and completeness. In contrast, our method avoids these issues and preserves both the fine details and the overall structural integrity of the reconstructed urban scene.

Table \ref{tab:mesh} presents a quantitative evaluation of reconstruction quality on the GauU-Scene dataset \cite{xiong2024gauuscene}, using precision, recall, and F1-score metrics at the distance threshold $\tau=0.025$. Our proposed approach demonstrates superior performance across all evaluated scenes, significantly outperforming other single GPU-based Gaussian Splatting (GS) methods in both precision and completeness metrics. Specifically, our method consistently achieves the highest or second-highest scores, reflecting robust reconstruction accuracy and completeness.

\begin{figure*}[!ht]
    \centering
    \includegraphics[width=0.9\linewidth]{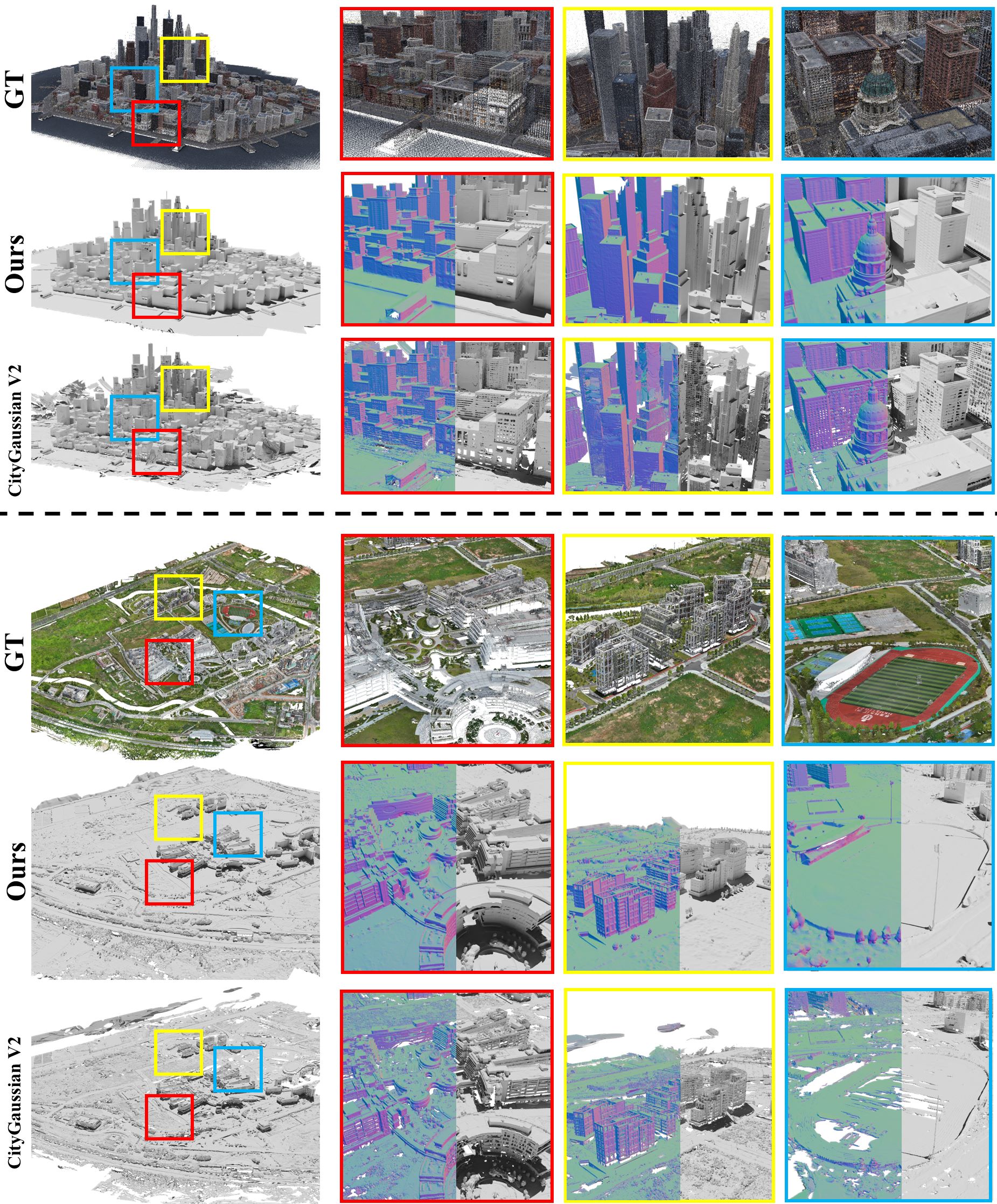}
    \caption{\textbf{Qualitative comparison between ULSR-GS and CityGaussianV2 \cite{liu2024citygaussianv2efficientgeometricallyaccurate} on the Matrix City dataset \cite{li2023matrixcity} \textit{Small City} scene and the cumtom collected dataset \textit{Scene1}}. Our method is superior to CityGaussianV2 \cite{liu2024citygaussianv2efficientgeometricallyaccurate} in terms of building integrity and detail preservation.}
    \label{fig:city_gaussian_comparison}
    
\end{figure*}

Regarding computational efficiency, our method, running on four GPUs, achieves shorter training times compared to single-GPU methods GOF \cite{Yu2024GOF} and PGSR \cite{chen2024pgsr} at 60K iterations. Notably, GOF and PGSR suffer from OOM issues in specific complex scenes due to their similar densification strategies, limiting their applicability for large-scale reconstructions. This clearly highlights the efficacy and scalability of our partitioning strategy, which effectively mitigates GPU memory limitations and enables comprehensive reconstruction even in resource-constrained environments. Furthermore, detailed qualitative visual comparisons in Figure \ref{fig:mesh_comp} clearly illustrate the superior detail reconstruction capability of our method. The magnified detail results highlight significant improvements in geometric accuracy and fine detail representation compared to the baseline methods. 

\subsubsection{Comparison with multi-GPU-based GS Methods.}

\begin{figure*}[!ht]
    \centering
    \includegraphics[width=1\linewidth]{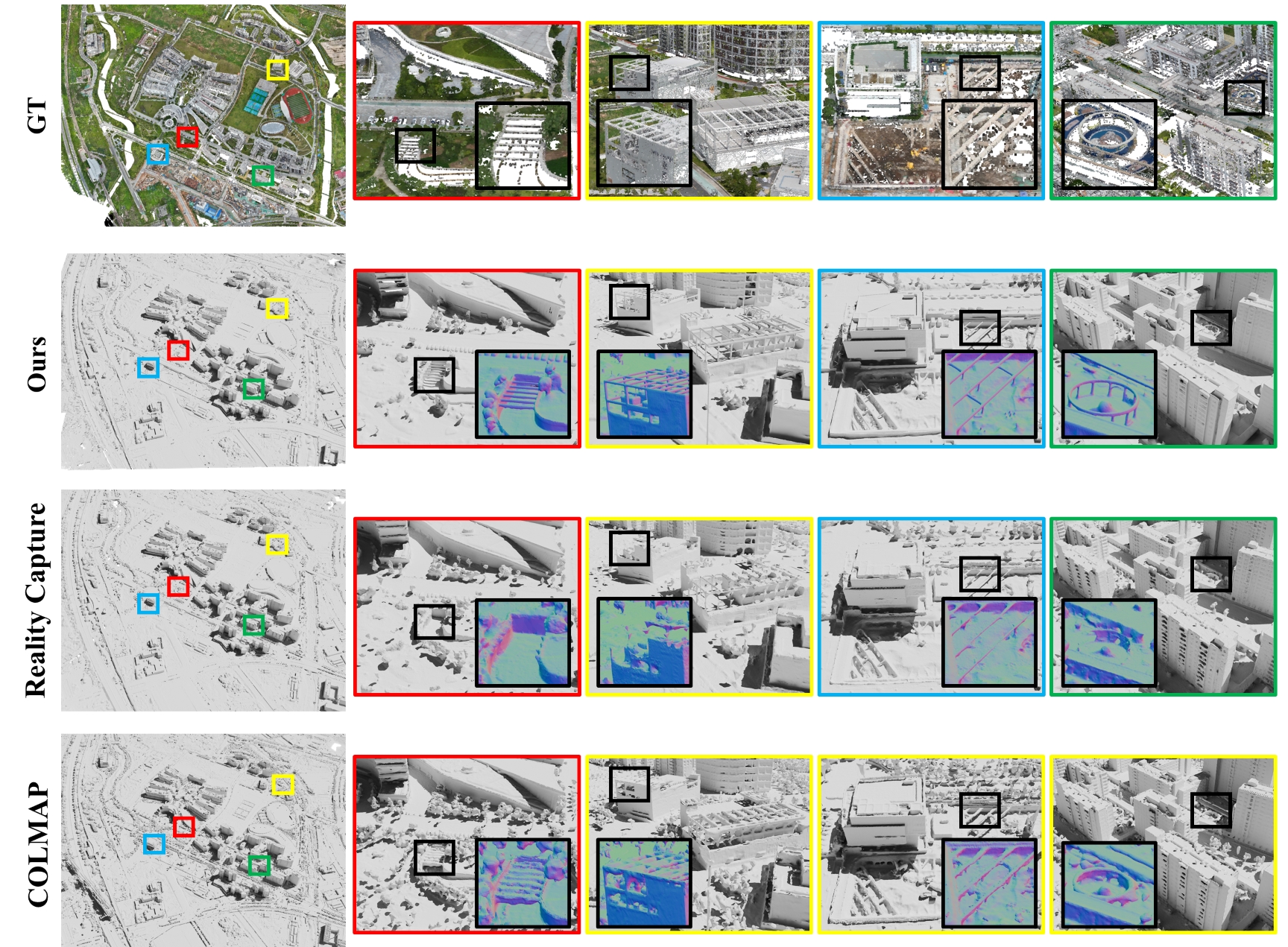}
    \caption{\textbf{Qualitative comparison between ULSR-GS and Reality Capture and COLMAP on the cumtom collected dataset \textit{Scene1}}. We use a image downsampling factor of 4$\times$ on every method, and use the highest setting of Reality Capture and COLMAP for fair comparison. It can be seen that with the same downsampling factor, ULSR-GS can better reconstruct the thin geometric structure of the scene. }
    \label{fig:ours_rc}
\end{figure*}

\begin{figure*}[!hb]
    \centering
    \includegraphics[width=1\linewidth]{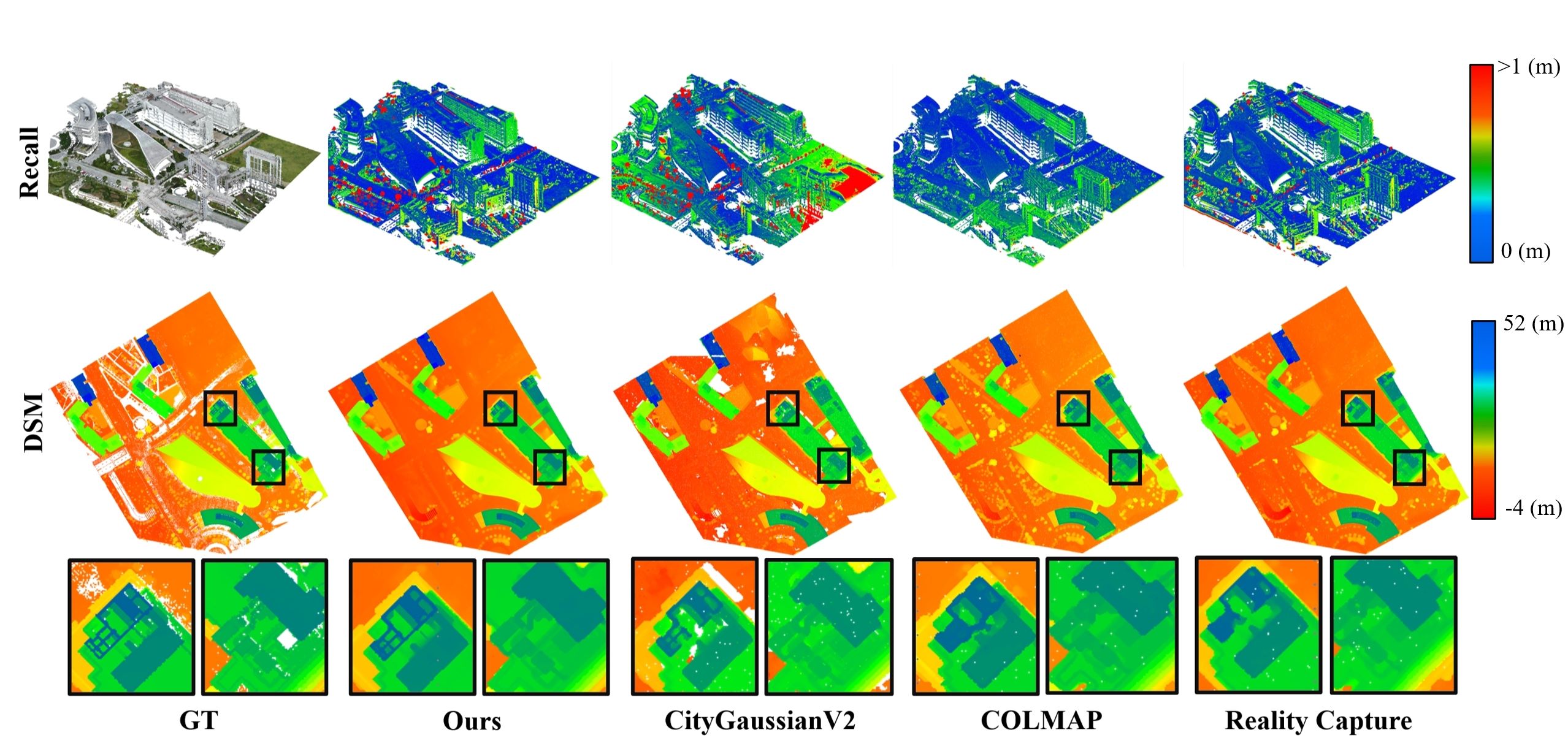}
    \caption{\textbf{Qualitative evaluation of the Recall metric and the generated DSM.}}
    \label{fig:dem}
\end{figure*}

\begin{figure*}[!ht]
    \centering
    \includegraphics[width=1\linewidth]{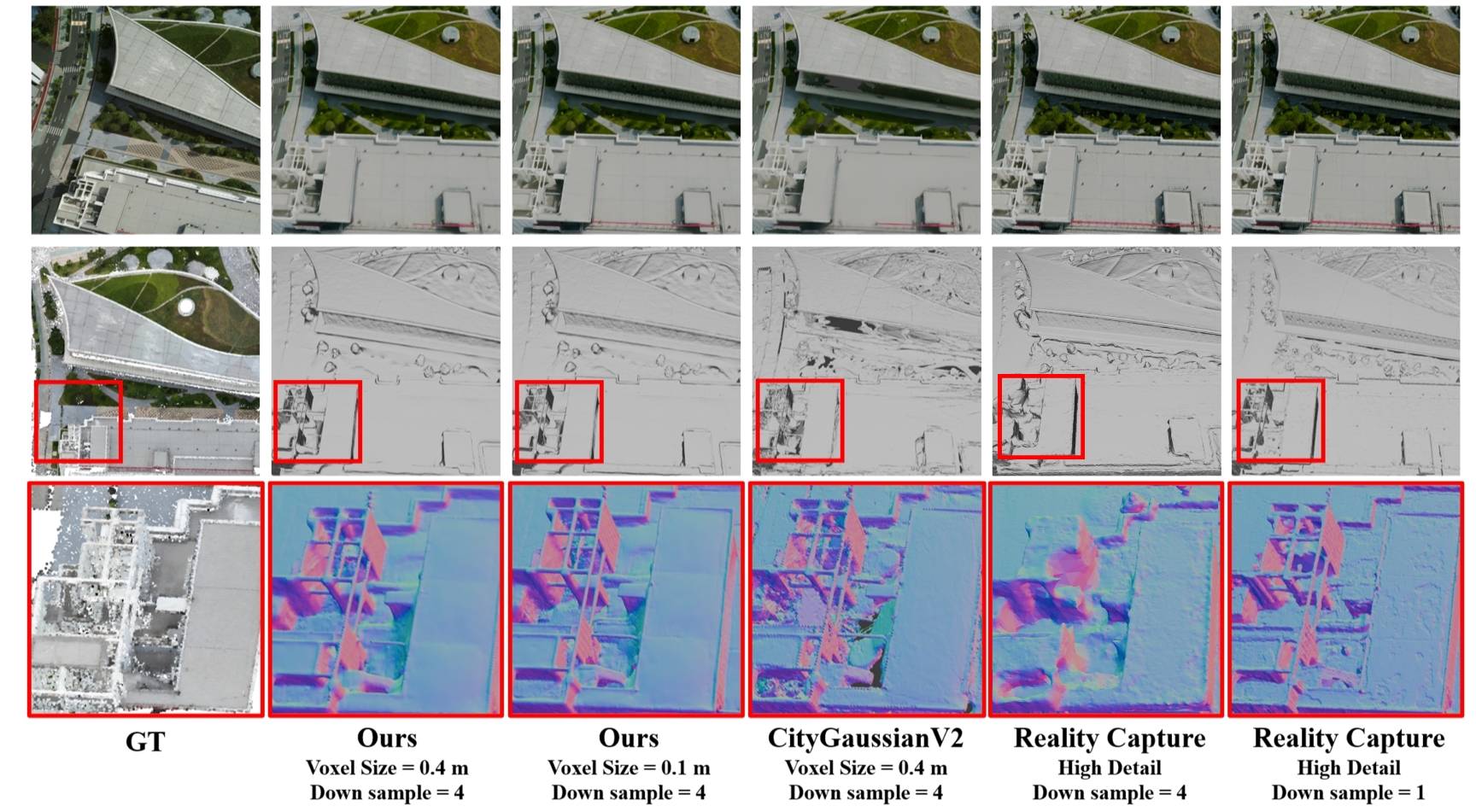}
    \caption{\textbf{Qualitative evaluation of different parameter settings.} From left to right: GT LiDAR, ULSR-GS with TSDF-fusion at 4x downsampling and 0.4m voxel size, ULSR-GS with TSDF-fusion at 4x downsampling and 0.1m voxel size, CityGaussianV2 with TSDF-fusion at 4x downsampling and 0.4m voxel size, RealityCapture with 4x downsampling and reconstruction quality set to “High Detail”, and RealityCapture with 1x downsampling and reconstruction quality set to “High Detail”.}
    \label{fig:voxel}
\end{figure*}

Table \ref{tab:mesh1} presents a quantitative evaluation comparing our proposed method with existing multi-GPU-based GS methods on the Matrix City dataset \cite{li2023matrixcity}, and two additional custom-collected datasets contain nearly 10,000 images (Scene 1 and Scene 2). Precision, recall, and F1-scores are reported at two different distance thresholds ($\tau=1.0m$ and $\tau=0.5m$). 

Our method consistently achieves the best or second-best results across metrics, underscoring the effectiveness and superiority of our per-point partitioning strategy in overcoming scalability bottlenecks, enabling efficient and accurate reconstruction of complex urban scenes.
CityGaussianV2 \cite{liu2024citygaussianv2efficientgeometricallyaccurate} requires initial coarse training on a single GPU followed by partitioning for multi-GPU fine-tuning. However, this training paradigm becomes infeasible for datasets comprising nearly 10,000 images, such as Scene1 and Scene2, where the initial sparse point cloud generated by COLMAP exceeds 30 million points, surpassing the memory limitations of a single RTX 4090 GPU with 24GB VRAM.
To address this challenge, we applied our proposed per-point partitioning strategy, dividing these two extensive scenes into smaller, manageable 2×2 sub-regions to apply the standard CityGaussianV2 training strategy. Consequently, as shown in Table \ref{tab:mesh1}, this adaptation significantly improves the performance of CityGaussianV2  on the larger scenes. 

In the qualitative comparison shown in Figure \ref{fig:city_gaussian_comparison}, CityGaussianV2 generates meshes with relatively sharp edges; however, significant holes emerge in smooth and low-texture regions such as roads and ground surfaces, negatively impacting its recall metrics and consequently resulting in lower F1-scores. Additionally, reflective surfaces, such as building glass facades, are inadequately reconstructed by CityGaussianV2, leading to incomplete structural representations. In contrast, our method successfully achieves a balanced reconstruction in both geometric detail and completeness, consistently outperforming CityGaussianV2 in both qualitative visual comparisons and quantitative metrics.

\subsubsection{Comparison with MVS-based Methods.}

\begin{table}[ht]
   \footnotesize
    \setlength\tabcolsep{1pt}
    \renewcommand\arraystretch{1.2}
    \centering 
    \caption{\textbf{Quantitative between MVS-based and GS-based Meshes.} We report precision, recall, and F-1 score at distance threshold $\tau = 0.5 m$. The \textbf{best}, \underline{second} results are highlighted. "OOM" denotes no results due to VRAM/RAM out of memory during training or TSDF-fusion. "ds" denotes image down sample factor and "v" denotes voxel size. We also report the total time of obtion final mesh results (train + extraction). } 
    \resizebox{\linewidth}{!}{
    \begin{tabular}{l|ccc|cc} % 16列
        \toprule
        Metrics  & Pre.$\uparrow$ & Rec.$\uparrow$ & F1$\uparrow$&   Face (M)   & Time (h) $\downarrow$ \\
        \midrule
        
        COLMAP (ds=4)& {0.630} &  {0.798} &  {0.704} & 372.3 &  29.1   \\

        COLMAP (ds=1)& \textbf{{0.707}} &  \underline{0.814} &  \underline{0.756} & 627.9  & 69.2 \\

        Reality Capture (ds=4)& {0.629} &  {0.794} &  {0.702} & \textbf{335.6} &  17.7 \\

        Reality Capture (ds=1)& \underline{0.703} &  \textbf{{0.832}} &  {\textbf{0.762}} & 1021.2 & 61.4\\

        \midrule
        2DGS (ds=4) & OOM & OOM & OOM & OOM & OOM   \\

        CityGSV2 (ds=4, v=0.4m) &  {0.646} & {0.753} & {0.695} & \underline{357.2} & \textbf{16.1}  \\

        CityGSV2 (ds=4, v=0.1m) & OOM & OOM  & OOM & OOM & OOM    \\

        Ours (ds=4, v=0.4m) &      {0.632} &  {0.787} &  {0.701} & \textbf{335.6} & \underline{16.7}  \\

        Ours (ds=4, v=0.1m) &      {0.667} &  {0.799} &  {0.727} & 983.5 & 19.8  \\

        \bottomrule
    \end{tabular} }
    
    \label{tab:mesh_mvs}  
\end{table}

From Table \ref{tab:mesh_mvs}, we compare our method against traditional MVS-based approaches (e.g., COLMAP, Reality Capture) and GS-based methods (e.g., 2DGS, CityGSV2). Our approach demonstrates notable advantages in terms of reconstruction efficiency, memory efficiency, and structural detail preservation. As shown in Table \ref{tab:mesh_mvs}, our approach achieves a competitive F1 score (0.701 at ds=4, v=0.4m) while maintaining a significantly lower computational cost (16.7 hours) compared to MVS-based methods like Reality Capture, which requires ds=1 to achieve better detail (F1: 0.762, 61.4 hours). Qualitative results in Figure \ref{fig:ours_rc} further illustrate that our method preserves fine structures, such as pipes and complex architectural elements, even at ds=4, whereas Reality Capture produces smoother, less detailed meshes. Additionally, recall visualizations in Figure \ref{fig:dem} and \ref{fig:voxel} highlight that our approach generates detailed DSMs at ds=4, whereas MVS-based methods require ds=1 to achieve comparable results, leading to increased computational overhead. Unlike other GS-based methods that suffer from memory limitations at fine voxel resolutions (v=0.1m), our method remains computationally feasible while achieving a high F1 score. These results emphasize the efficiency of our approach in large-scale urban reconstruction, providing high-quality 3D meshes with reduced processing time and memory requirements.

\subsection{Ablation Study.} 
\begin{table}[h!]
    \centering
    \setlength{\tabcolsep}{2pt}
    
    \scriptsize
    \caption{\textbf{Geometric Consistency Constraints Ablation Studies.} We conducte ablation experiments on the GauU-Scene  dataset \cite{xiong2024gauuscene} in 4$  \times$4 cells. We reported metrics including Precision, Recall, F-1 Score. "-" denotes no data due to GPU out of memory.}
    \begin{tabular}{l|ccc}
        \toprule
        & Precision. $\uparrow$ & Reccall. $\uparrow$ & F-1 Score $\uparrow$   \\
        \midrule
        A.  w/o $E_{depth}$ \& $E_{normal}$  & 0.659 & 0.661 & 0.660 \\
        B. w/o $E_{depth}$ & 0.663 & 0.665 & 0.664  \\
        C. w/o $E_{normal}$ & 0.652 & 0.651 & 0.651  \\
        D. w/o densify & 0.678 & 0.685 & 0.682  \\
        E. w densify \& w/o window mask & - & - & -  \\
        \midrule
        F. Full   & 0.681 & 0.689 & 0.685  \\
        \bottomrule
    \end{tabular}
    
    \label{table:abl_loss}
\end{table}

\begin{figure}[h!]
    \centering
    \includegraphics[width=1\linewidth]{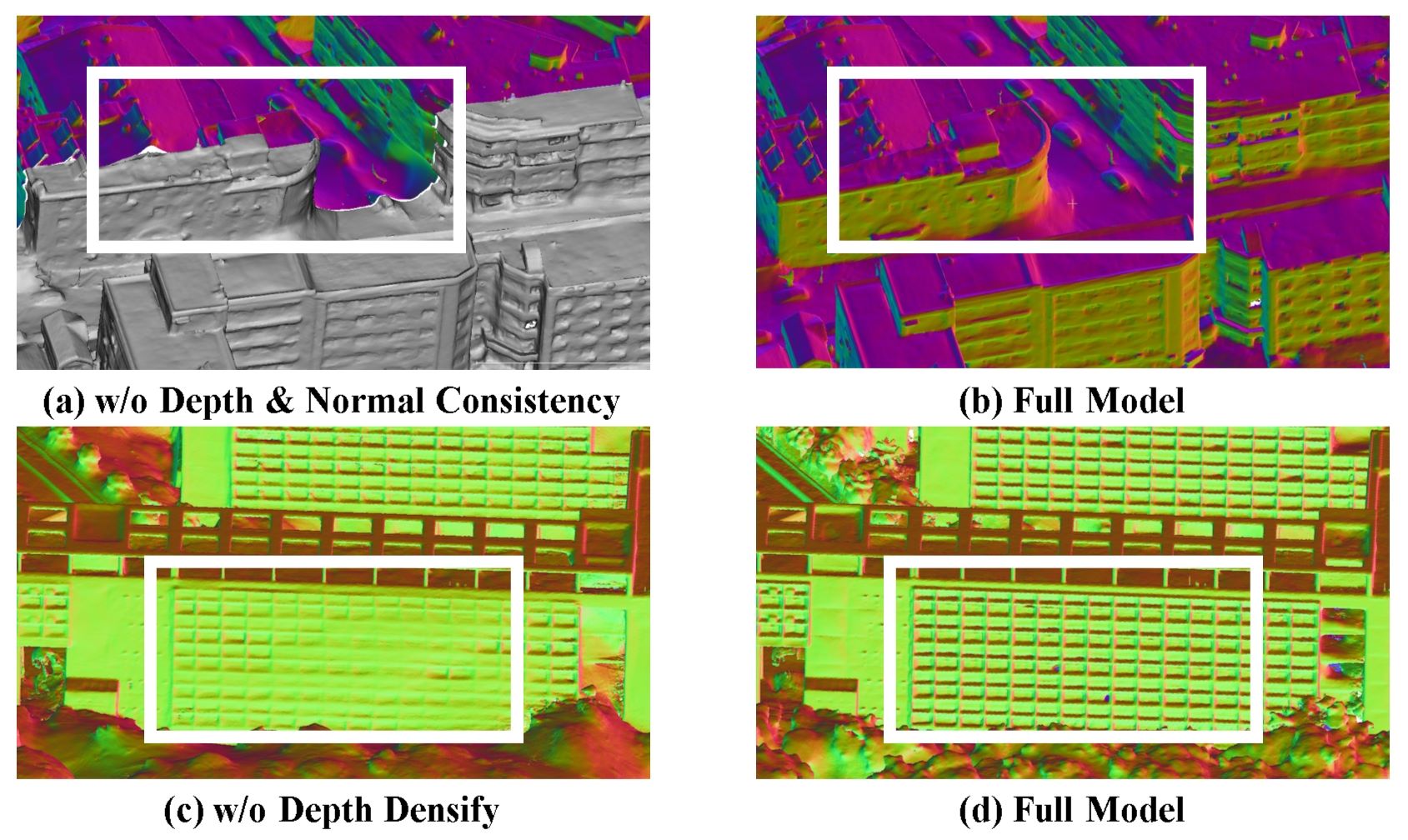}
    
    \caption{\textbf{Qualitative comparison on the geometric consistency constraints ablation studies.} Without $E_{depth}$ \& $E_{normal}$ edges of each region fail to stitch seamlessly; Without adaptive densify results in excessively smooth mesh extraction outcomes.  }
    \label{fig:loss}
\end{figure}

\begin{figure*}[!hbp]
    \centering
    \includegraphics[width=1\linewidth]{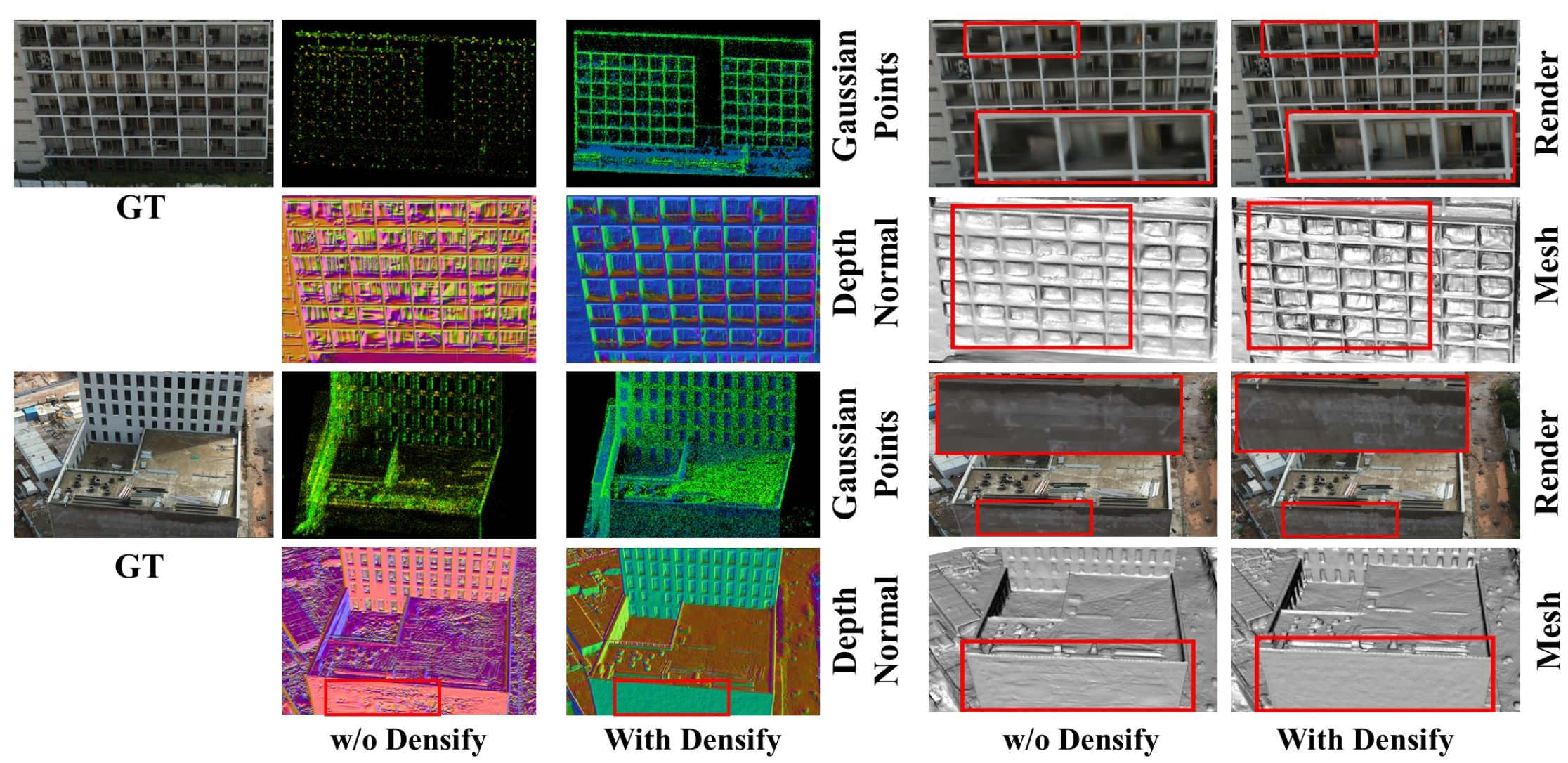}
    
    \caption{\textbf{Qualitative comparison on the adapted depth densification ablation studies.} Without adaptive densification (e.g. vanilla 2dgs greadient-based densification) results in excessively smooth mesh extraction outcomes and loss rendering quality.}
    \label{fig:/densify}
\end{figure*}

Table \ref{table:abl_loss} presents the results of our ablation experiments, which quantitatively demonstrate the impact of removing geometric consistency constraints on reconstruction quality. Specifically, we observe that the absence of these constraints leads to a notable decline in Precision, Recall, and F1 Score, indicating that geometric consistency is essential for both accuracy and completeness.
\subsubsection{Impact of Removing Individual Constraints
}

When both multi-view depth and normal consistency constraints are removed, it exhibits a noticeable performance decline, reinforcing the importance of geometric constraints in high-quality reconstruction. Interestingly, the F1 score for this condition is slightly higher than that of removing normal consistency alone (Condition C). This suggests that when only normal consistency is absent, the retained depth consistency may introduce conflicting geometry that worsens reconstruction errors, whereas removing both constraints forces the model to rely solely on single image-based information, leading to a different but somewhat more stable degradation. Figure \ref{fig:loss} (a) and (b) further highlight the dominant role of multi-view depth and normal consistency in preventing surface misalignment and ensuring smooth reconstructions.

When only the multi-view depth consistency is removed while retaining normal consistency, the moderate decline in performance suggests that although the absence of depth consistency introduces minor geometric misalignment across views, the retained normal consistency still preserves local surface smoothness and alignment, preventing a severe degradation in quality. This result implies that while depth consistency helps refine the global structure, normal consistency alone can still enforce a degree of geometric regularization, maintaining an acceptable level of reconstruction fidelity.
When multi-view normal consistency is removed but depth consistency is retained, we observe a substantial drop in performance. This suggests that normal consistency plays a more critical role in maintaining surface quality than depth consistency alone. Without normal consistency, individual surfaces reconstructed from different views may exhibit inconsistent orientations, leading to visible artifacts, non-smooth transitions, and misaligned surface patches. 

\subsubsection{Impact of Densification Strategies.}
When the densification process is removed, Precision and Recall drop slightly to 0.678 and 0.685, respectively, with an F1 Score of 0.682. However, the qualitative comparison in Figure \ref{fig:loss} and \ref{fig:/densify} suggests that densification primarily refines the reconstruction rather than fundamentally altering its structure. The results indicate that while densification improves point distribution and mesh completeness, the geometric consistency constraints (depth and normal consistency) have a more significant impact on the overall reconstruction quality. When removing the window mask, ULSR-GS runs out of memory (OOM), preventing evaluation. This suggests that removing the window mask leads to an excessive increase in Gaussian points, which significantly inflates the computational burden. The failure to complete the reconstruction in this setting highlights the importance of an adaptive sparsification mechanism, which prevents unnecessary computational overhead while preserving reconstruction quality.

\subsubsection{Impact of Partition Strategies }
\textit{(a) Geometry Reconstruction}

Table \ref{table:abl_pp} presents comprehensive comparative experiments analyzing different partitioning strategies. In the absence of optimized pair selection (as demonstrated in Table \ref{table:abl_pp}(A)), a significant number of redundant images are included in training for each region. This redundancy not only substantially increases computational overhead but also negatively impacts the quality and precision of the extracted meshes. These redundant images typically originate from inaccuracies inherent to SfM computations \cite{schoenberger2016colmap}, which fail to effectively filter images based solely on geometric consistency.

To further illustrate the robustness and effectiveness of our proposed partitioning strategy, we conducted experiments comparing our strategy against the image location-based partitioning method employed by Vast Gaussian \cite{vastgaussian}. Specifically, we evaluated both ULSR-GS and the baseline method 2DGS \cite{huang20242dgs} under different partitioning conditions (Table \ref{table:abl_pp}(B), (C), and (D)). Our results demonstrate a clear superiority of our method, particularly in large-scale surface reconstruction scenarios. 

\begin{figure*}[hb!]
    \centering
    \includegraphics[width=0.7\linewidth]{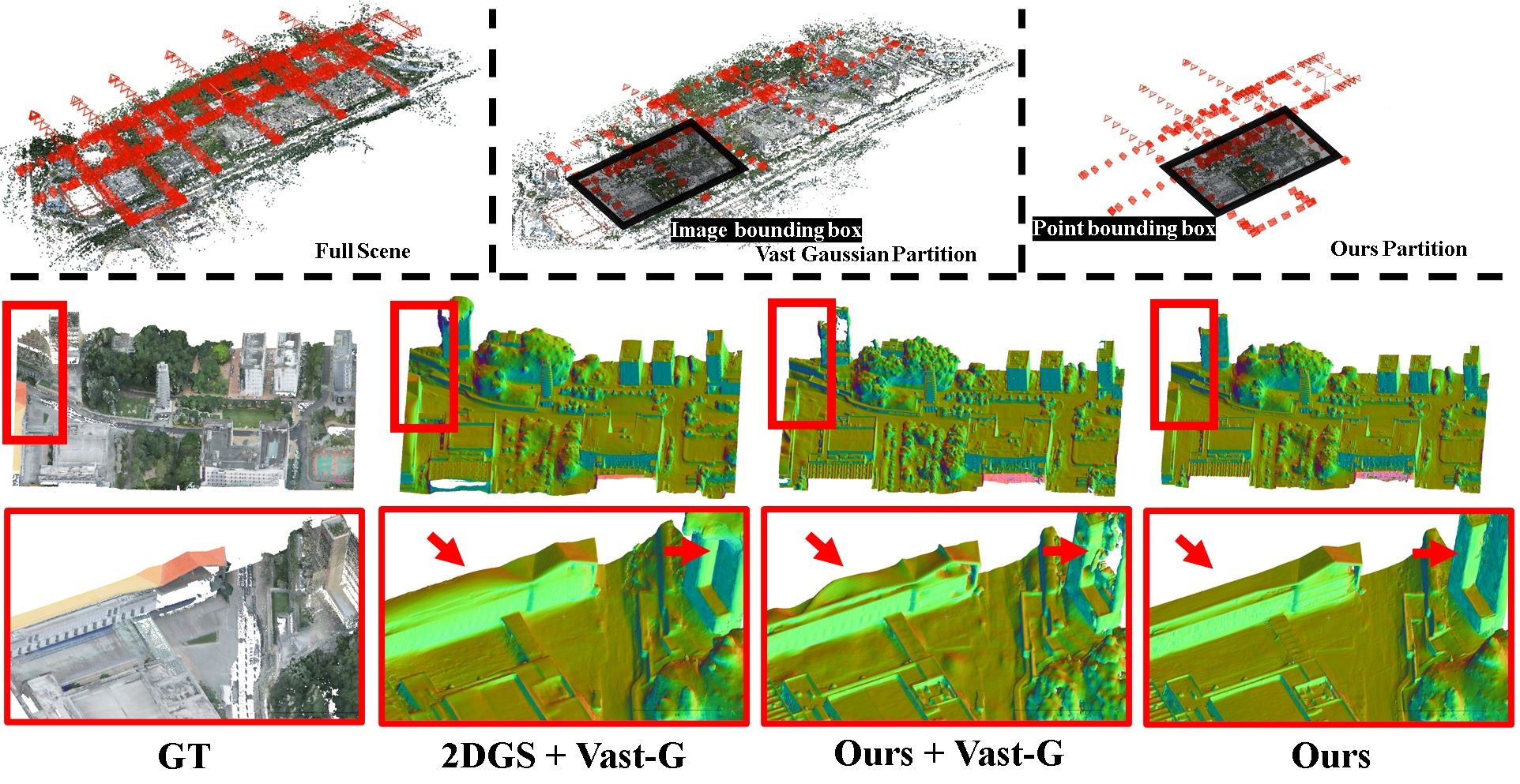}
    \caption{\textbf{Qualitative comparison of partition strategies on a penta-cam dataset.} The \textbf{\textit{black}} bounding box is the same partition area. The image partition boundary for VastGaussian and and the point cloud partition boundary for ULSR-GS ars set to the same size according to the black boundary. Below are the meshes after clipping to the box size.}
    \label{fig:lower_comp}
\end{figure*}

\begin{figure*}[hb!]
    \centering
    \includegraphics[width=0.7\linewidth]{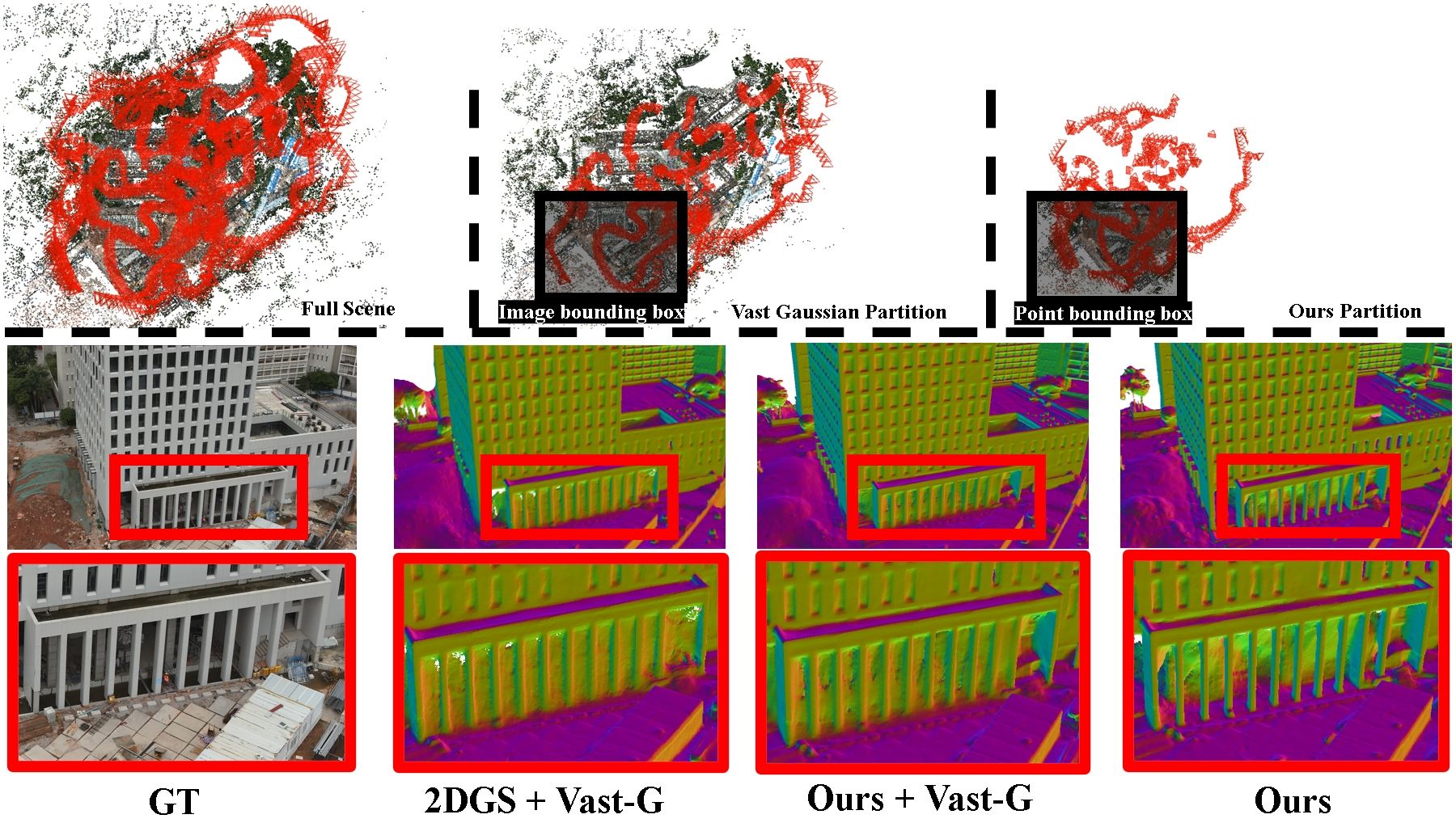}
    \caption{\textbf{Qualitative comparison of partition strategies on a close-range dataset.} The \textbf{\textit{black}} bounding box is the same partition area. The image partition boundary for VastGaussian and and the point cloud partition boundary for ULSR-GS ars set to the same size according to the black boundary. Below are the meshes after clipping to the box size.}
    \label{fig:camp_pp}
\end{figure*}

Notably, Figure \ref{fig:camp_pp} visually illustrates the effectiveness of our partitioning strategy using a more challenging close-range oblique photogrammetry dataset featuring intricate and specific UAV flight paths \cite{UrbanScene3D}. In this dataset, our approach successfully identifies and selects images highly relevant to the reconstruction region, significantly enhancing local detail preservation and accuracy. Conversely, the partitioning strategy of Vast Gaussian, which relies strictly on the projection of camera positions onto a 2D plane, selects images from a broader area without adequate precision. Consequently, critical images positioned outside of a predefined boundary, yet crucial for reconstructing complex structures, are erroneously excluded. Simultaneously, the inclusion of unrelated, distant images introduces noise, ultimately deteriorating the reconstruction quality. Additionally, on datasets collected using multi-directional Penta-Cam setups in Figure \ref{fig:lower_comp}, where camera paths and orientations are highly regular, our point-to-photo partitioning approach consistently outperforms the camera projection-based method.

 \begin{figure*}[!t]
     \centering
     \includegraphics[width=0.7\linewidth]{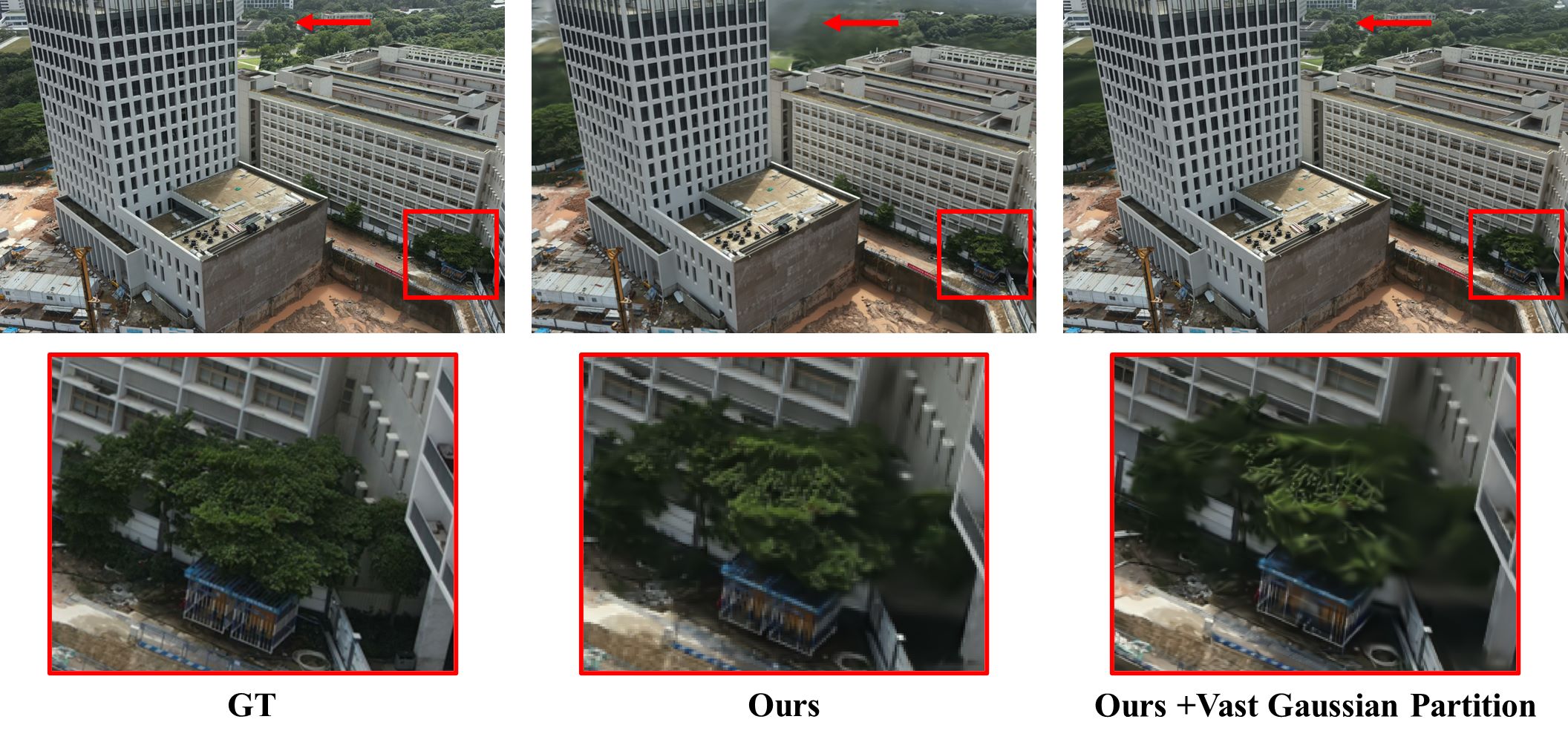}
     \caption{\textbf{Qualitative comparison of partition strategies.} In close-range photography tasks, a large number of background areas will appear. Our partitioning method emphasizes high-quality rendering of the reconstructed subject and ignores the division of background areas.}
     \label{fig:artsci1}
 \end{figure*}

% Since UrbanScene 3D \cite{UrbanScene3D} does not provide a reference ground truth point cloud, we perform a qualitative analysis. And Figures \ref{fig:lower_comp} and \ref{fig:camp_pp} qualitatively demonstrate comparisons on both five-directional oblique photogrammetry datasets and close-range oblique photogrammetry datasets.  Our partitioning strategy effectively identifies and selects the most critical images for the reconstruction regions, significantly enhancing the detail and accuracy of the extracted meshes. 

\textit{(b) Novel view synthesis.}

Table \ref{tab:partition_comp} quantitatively compares different partitioning methods, specifically our point-based partitioning approach and the image-based method employed by Vast Gaussian \cite{vastgaussian}. The results indicate that the Vast Gaussian's \cite{vastgaussian}] partitioning strategy is more suitable for rendering tasks. This is because our method prioritizes the reconstruction of the scene's primary structure by excluding the distant sparse SfM points during sub-region partitioning. Figure \ref{fig:artsci1} provides a qualitative comparison to further illustrate this difference. When training ULSR-GS using different partitioning strategies, our partition method shows a trade-off: while it loses performance in reconstructing the background, it preserves significantly more detail in the reconstruction of the primary scene components.

% When the number of images in the scene (\textit{LFLS}, \textit{SZIIT} (see Figure \ref{fig:sztu}), and \textit{SZTU}) exceeds 1000, the reconstruction quality indicators of 2DGS \cite{huang20242dgs}, PGSR \cite{chen2024pgsr} and GOF \cite{Yu2024GOF} are far behind us. 

\begin{table}[!hbp]
    \setlength\tabcolsep{1pt}
    \renewcommand\arraystretch{1.2}
    \centering 
    \caption{\textbf{Partition Strategies Comparison.} We conduct experiments on the GauU-Scene  dataset \cite{xiong2024gauuscene} scene in 4$\times$4 cells. We reported metrics including Precision, Recall, and F-1 Score, Average number of images per cell, and Total training time. VastG $^\nmid$ means the unofficial reproduction partition method of Vast Gaussian \cite{vastgaussian}. }
    \resizebox{\linewidth}{!}{
    \begin{tabular}{l|ccc|cc}
        \toprule
       \textbf{GauU-Scene} (Avg. images: 957)  & Pre. $\uparrow$ & Rec. $\uparrow$ & F1 $\uparrow$ & 
        Avg. I. &Time $\downarrow$   \\
        \midrule
        A. Ours w/o View Selection & 0.699 & 0.707 & 0.702& 437& 4.5 h  \\
        B. Ours + Vast-G $^\nmid$   & 0.685 & 0.700 & 0.692& 359 & 3.5 h  \\
        C. 2DGS + Ours   & 0.677 & 0.671 & 0.674 & 366& 3.0 h  \\
        D. 2DGS + Vast-G $^\nmid$  & 0.663 & 0.669 & 0.667 &  359& 2.9 h  \\
        \midrule
        E. 2DGS-60K (1 GPU)  & 0.573 &  0.631& 0.600& 957 & 2.1 h \\
        F. ULSR-GS (4 GPU)  & 0.713 & 0.725 & 0.717 & 366
 & 3.7 h \\
        \bottomrule
    \end{tabular}}
    
    \label{table:abl_pp}
\end{table}

% \begin{figure*}[!hbp]
%     \centering
%     \includegraphics[width=0.8\linewidth]{images/boundary.jpg}
%     \caption{\textbf{Qualitative comparison of partition strategies at scene edges.}  We train our method on our partition strategy and Vast Gaussian's \cite{vastgaussian} in $8\times8$ cells. From top to bottom: example scene edge, example GT images at the scene edge, ULSR-GS on our partition strategy, and ULSR-GS on Vast Gaussian's \cite{vastgaussian} partition strategy.}
%     \label{fig:boundary}
% \end{figure*}

\begin{table}[!hb]
\setlength\tabcolsep{1pt}
    \renewcommand\arraystretch{1.2}
    \centering 
    \resizebox{\linewidth}{!}{
\begin{tabular}{c|ccc|ccc|ccc}
\toprule
Data Type       & \multicolumn{3}{c|}{Five-directional} & \multicolumn{6}{c}{Close-range}            \\
\midrule
Scene           & \multicolumn{3}{c|}{\textbf{CUHK\_LOWER \cite{xiong2024gauuscene}}}     & \multicolumn{3}{c|}{\textbf{Artsci \cite{UrbanScene3D}}} & \multicolumn{3}{c}{\textbf{Polytec \cite{UrbanScene3D}}} \\
\midrule
Metrics         & PSNR$\uparrow$       & SSIM $\uparrow$       & LPIPS $\downarrow$   & PSNR $\uparrow$    & SSIM $\uparrow$   & LPIPS  $\downarrow$   & PSNR $\uparrow$     & SSIM $\uparrow$   & LPIPS   $\downarrow$  \\
\midrule
2DGS            &     23.85       &      0.711       &      0.382      & 25.42  & 0.763  & 0.346  &     24.96     &   0.849     &   0.236      \\
2DGS+Vast-G &  \underline  {24.98}        &    0.801     &      0.247         &   26.98  &   0.845  &   0.299  &      26.31     &    0.860     &    0.221         \\

2DGS+Ours Partition &      24.87       &    0.789        &     0.250       &   26.72  & 0.826  & 0.312  &     26.02     &   0.855     &   0.229         \\
Ours+Vast-G &  \textbf  { 25.21 }      &   \underline   {0.805 }     &   \textbf   {0.234 }     & \textbf {27.14}  & \textbf {0.883 } & \textbf  {0.280}  &  \textbf    {27.11}     &  \textbf  {0.869}     & \textbf  { 0.207 }        \\
\midrule
Ours            &   \textbf  {25.21}       &  \textbf   {0.810 }      &   \underline  {0.239}       &  \underline  {27.01}  &\underline  {0.851}  & \underline 0.289  &     \underline {26.97}     & \underline   {0.861}     & \underline  { 0.213 }        \\
\bottomrule
\end{tabular}}
\caption{\textbf{Additional partition comparison.} We conduct comparative experiments on different types of oblique photography datasets. Our partition strategy overemphasizes the reconstruction of the main area and deletes the background area with only a few reconstructed images, resulting in a decrease in the overall rendering quality.}
\label{tab:partition_comp}
\end{table}

\section{Discussion}
\subsection{Real World Application: Integration with USV Bathymetry model}

\begin{figure}[!h]
    \centering
    \includegraphics[width=1\linewidth]{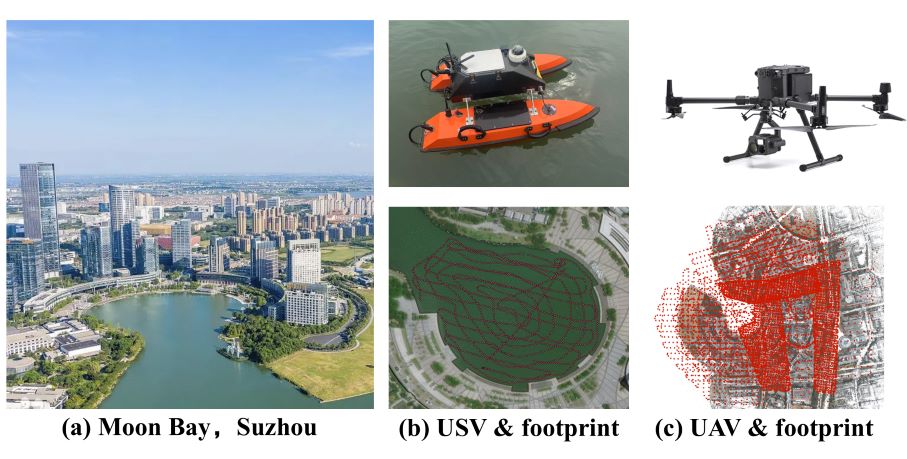}
    \caption{\textbf{Experimental Location.} (a) Bird-eye-view of the experimental location Moon Bay, Dushu Lake, Suzhou. (b) USV and its SBES mathymetry footprints. (c) UAV and its oblique photogrammetry footprints.}
    \label{fig:exploc}
\end{figure}

We now discuss the ability to integrateg the bathymetry digital elevation model (BDEM) obtained from a single-beam echo sounder (SBES) with the oblique photogrammetric models produced by ULSR-GS to generate a real urban surface model. 
As shown in Figure \ref{fig:exploc} (a), the experimental site for this study is located at Moon Bay (Scene 2), situated along the southern shore of Dushu Lake in Suzhou, China. Moon Bay is a scenic and ecologically rich area, but it is also surrounded by a dense cluster of urban buildings, which presents significant challenges for UAV-based photogrammetric reconstruction and the subsequent data fusion with USV-derived bathymetric models. 
% The key parameters of the equipments ars summarized in Table \ref{table:equp}.

% \begin{table}[!h]
%     \centering
%     \setlength{\tabcolsep}{4pt}
%     \scriptsize
%     \begin{tabular}{l|cc}
       
%         Product Mode       & HXF 260D SBES                  & M-300 RTK UAV   \\ \hline
% Spatial Range      & 1.0 - 200 m (vertical)            & 5000 m            \\
% Accuracy           & 1 cm ± 0.1\% of depth & ± 0.1 m (with RTK) \\
% Spatial Resolution & 1 m                   & 1 cm - 10 cm 
        
%     \end{tabular}
%     \caption{\textbf{Experimental Equipments.}}
%     \label{table:equp}
% \end{table}

\begin{figure}[htbp]
    \centering
    \includegraphics[width=1\linewidth]{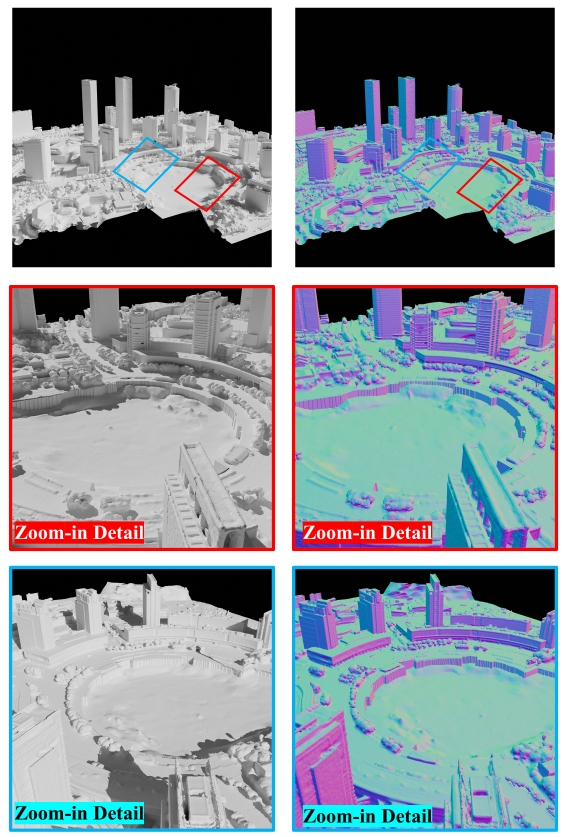}
    \caption{Zoomed-in detail of the fusion USV and UAV models. Left: Mesh; Right Normal.}
    \label{fig:moon_zoom_in}
\end{figure}

As shown in Figure \ref{fig:exploc} (b), the USV followed predefined paths across the water body, recording depth measurements at each point along its trajectory. The WGS-84 world coordinates were recorded with each measurement, providing a geospatial reference for each bathymetry point. Simultaneously, a UAV was deployed to capture oblique imagery of the surrounding landscape. Each image was georeferenced using the UAV's embedded GPS, recording the WGS-84 world coordinates at the time of capture. The initial data consists of bathymetry measurements at specific points. The bathymetric data collected by the USV is first processed using Geographic Information System (GIS) software, where the raw measurements are converted into a DEM through Kriging interpolation. The high-resolution DEM is converted into a 3D mesh using standard Triangulated Irregular Network (TIN) algorithms. As shown in Figure \ref{fig:moon_zoom_in}, the ULSR-GS-generated mesh and the UAV-derived mesh are then aligned and merged. The alignment process ensures that both meshes are represented in the same global coordinate system (WGS-84 Mercator projection), allowing for seamless integration.

% \begin{figure}[htbp!]
%     \centering
%     \includegraphics[width=1\linewidth]{images/artsci_com.jpg}
    
%     \caption{\textbf{Qualitative comparison of different partition strategy on close-range photogrammetry dataset.} We train 2DGS \cite{huang20242dgs} as baseline method on UrbanScene 3D \cite{UrbanScene3D} \textit{artsci\_corase} scene.}
%     \label{fig:artsci_com}
% \end{figure}

\subsection{Strengthsh and Limitations}

Our proposed ULSR-GS framework demonstrates notable advantages and specific application scenarios that clearly distinguish it from existing single-GPU \cite{guedon2023sugar,huang20242dgs,Yu2024GOF,zhang2024rade,chen2024pgsr}, multi-GPU \cite{liu2024citygaussianv2efficientgeometricallyaccurate,lin2024vastgaussian}, and MVS-based \cite{schoenberger2016colmap} methods. Compared to single-GPU GS approaches, ULSR-GS exhibits exceptional scalability and efficiency, making it particularly suitable for large-scale urban reconstruction tasks. The inherent memory constraints of single-GPU methods severely limit their capacity for handling extensive datasets; our multi-GPU strategy circumvents this limitation by employing an optimized point-to-photo partitioning technique driven by SfM. Consequently, ULSR-GS effectively reconstructs detailed urban scenes at scales previously unattainable with single-GPU methods \cite{guedon2023sugar,huang20242dgs,Yu2024GOF,zhang2024rade,chen2024pgsr}.

Furthermore, compared with other multi-GPU GS approaches \cite{liu2024citygaussianv2efficientgeometricallyaccurate,lin2024vastgaussian} that typically rely on image-centric partitioning, ULSR-GS provides greater adaptability and accuracy, especially beneficial in complex scenarios such as close-range and oblique photogrammetry. Our per-point partitioning ensures that each sub-region is reconstructed using its most relevant image set, thus addressing critical issues of uneven coverage and irregular camera paths encountered by traditional methods. These improvements are validated by quantitative metrics and qualitative evaluations across diverse urban scenes.
Moreover, relative to MVS-based methods \cite{schoenberger2016colmap}, ULSR-GS is particularly effective in reconstructing thin and complex architectural structures, which often present significant challenges for correspondence-based MVS techniques due to matching ambiguities or limited local texture.

Despite these strengths, ULSR-GS also exhibits several limitations: (1) As shown in Figure \ref{fig:dem}, the framework is less effective in vegetation-rich environments, where dense correspondence-based MVS techniques better capture the intricate and semi-transparent structures inherent in foliage; (2) the current GS-based training and rendering resolution limit (approximately 1.6K) is insufficient for applications demanding high-detail fidelity, such as high-resolution (2cm per pixel) oblique photogrammetry; and (3) the existing depth-projection-based densification strategy primarily enhances regions already adequately reconstructed, inadequately addressing areas that are initially sparse or poorly captured.

\section{Conclusion}
We present ULSR-GS, a framework dedicated to high-
fidelity surface extraction in ultra-large-scale scenes. Specifically, our partitioning approach combines with a multi-view selection strategy. Additionally, ULSR-GS employs a multi-view geometric consistency densification to enhance surface details. Experimental results demonstrate that ULSR-GS outperforms other SOTA GS-based works on large-scale benchmark datasets.
Our approach addresses key limitations of image-based partitioning methods \cite{vastgaussian,liu2024citygaussian,hierarchicalgaussians24} by ensuring finer granularity in mesh boundaries and prioritizing the reconstruction of key scene regions. This strategy enables us to extract meshes individually rather than merge sub-regions and extract the full scene. Unlike existing MVS-based approaches, our method enables the simultaneous generation of a high-quality radiance field for rendering and a detailed mesh representation for reconstruction, all within a single training pipeline.  Also, comparison with traditional MVS-based methods demonstrates ULSR-GS drastically reduces processing time while maintaining fine structural details. These advantages make ULSR-GS a promising choice for oblique photogrammetry, offering an efficient and scalable solution for real-world applications.

For future work, we identify three key directions to enhance our framework. First, improving depth computation  will be crucial for refining surface extraction and reducing artifacts in complex environments. Second, developing highly efficient 4K-resolution rendering and surface extraction will address the current limitations of GS-based methods in photogrammetric applications. Finally, aerial and terrestrial photogrammetry are combined to complete the detailed reconstruction of the building facade structure while solving the surface reconstruction of the translucent glass building.

\section{Acknowledgements}
This work was supported by the Suzhou Municipal Key Laboratory for Intelligent Virtual Engineering (SZS2022004), the XJTLU AI University Research Centre, the Jiangsu Province Engineering Research Centre of Data Science and Cognitive Computation at XJTLU, and the SIP AI Innovation Platform (YZCXPT2022103).

\bibliographystyle{cas-model2-names}

\bibliography{main}

 \clearpage
% \section{Additional Results}

\begin{figure*}[!hb]
    \centering
    \includegraphics[width=1\linewidth]{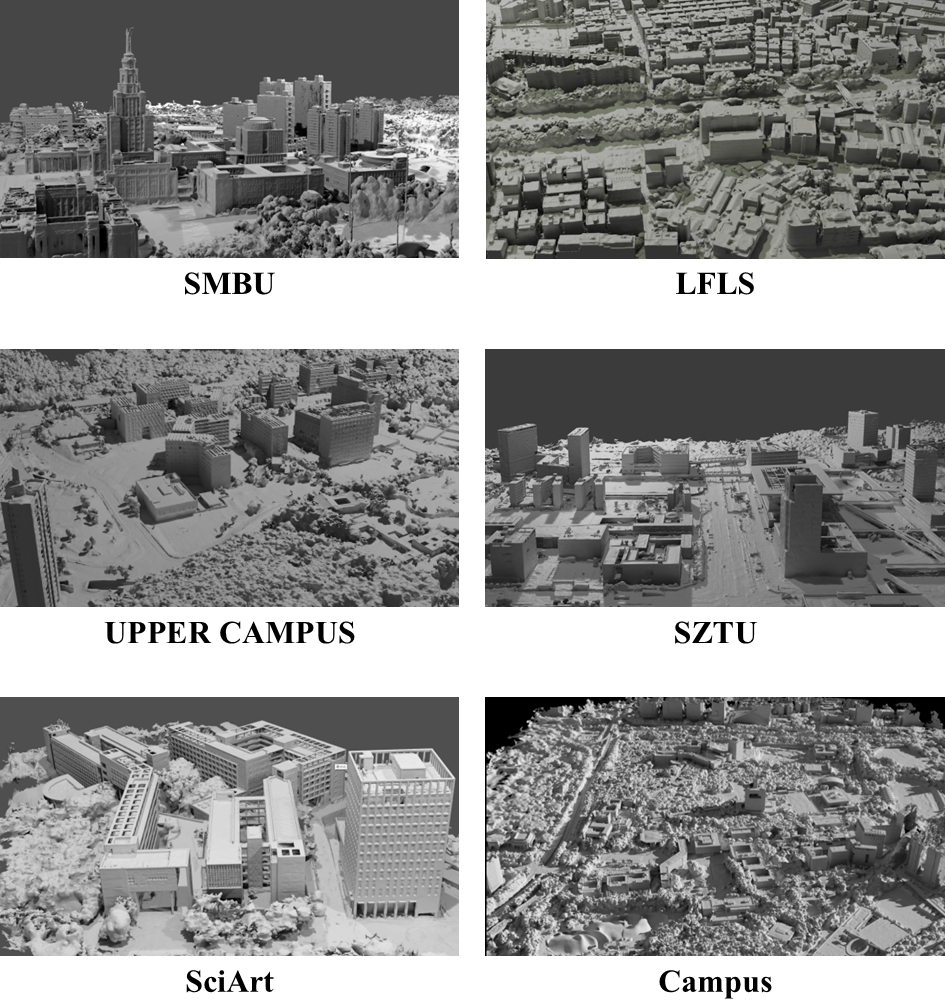}
    \caption{\textbf{Mesh Results on GauU-Scene \cite{xiong2024gauuscene} and UrbanScene 3D \cite{UrbanScene3D} Datasets.}}
    \label{fig:more_results}
\end{figure*}

\begin{figure*}
    \centering
    \includegraphics[width=1\linewidth]{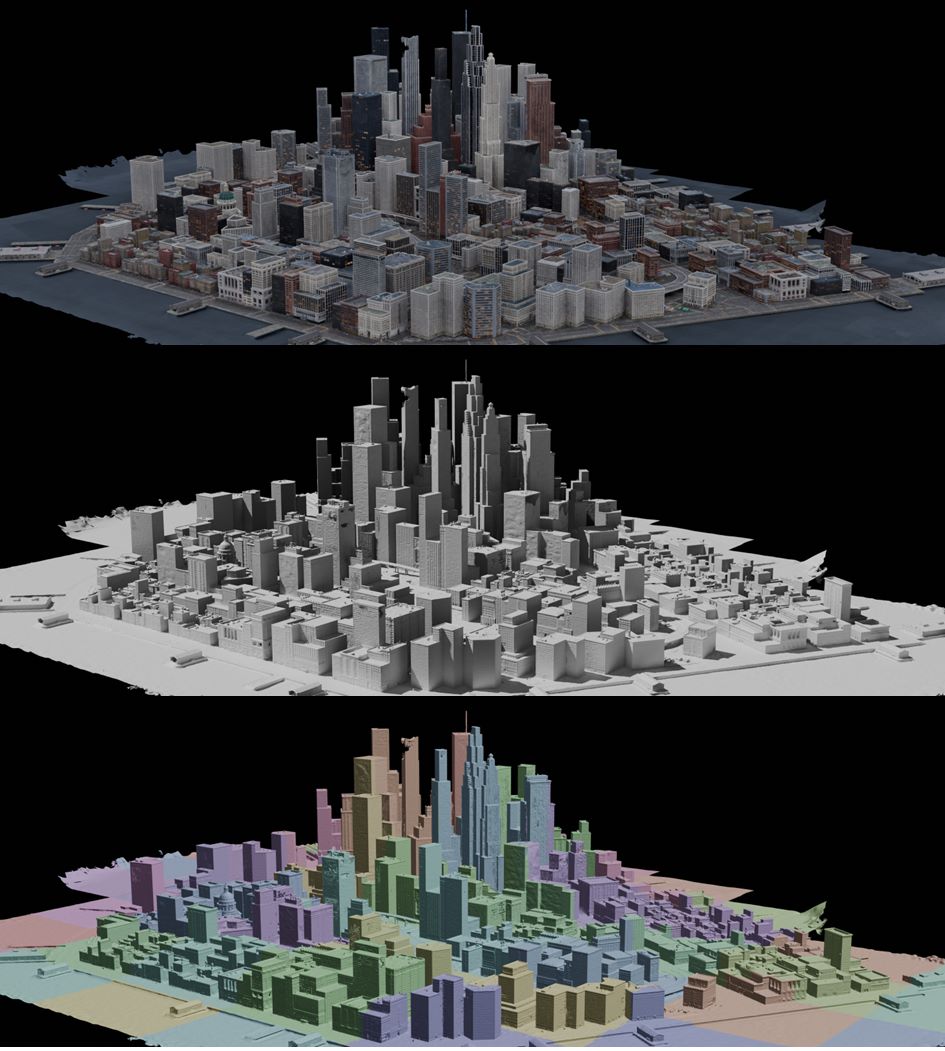}
    \caption{\textbf{Mesh Results on MatrixCity \cite{li2023matrixcity} Dataset.}}
    \label{fig:MC}
\end{figure*}

\begin{figure*}
    \centering
    \includegraphics[width=0.8\linewidth]{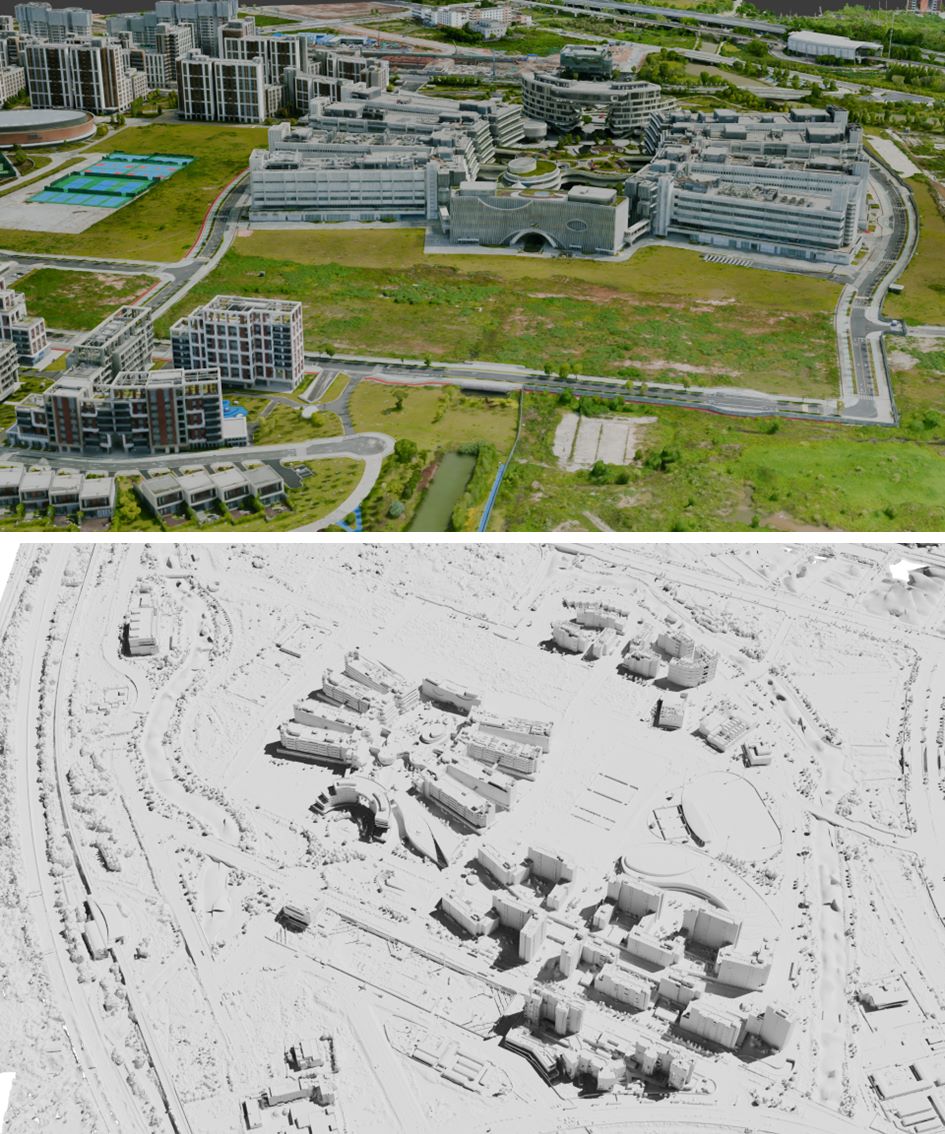}
    \caption{\textbf{Mesh Results on Scene 1.}}
    \label{fig:scane1}
\end{figure*}

\begin{figure*}
    \centering
    \includegraphics[width=0.8\linewidth]{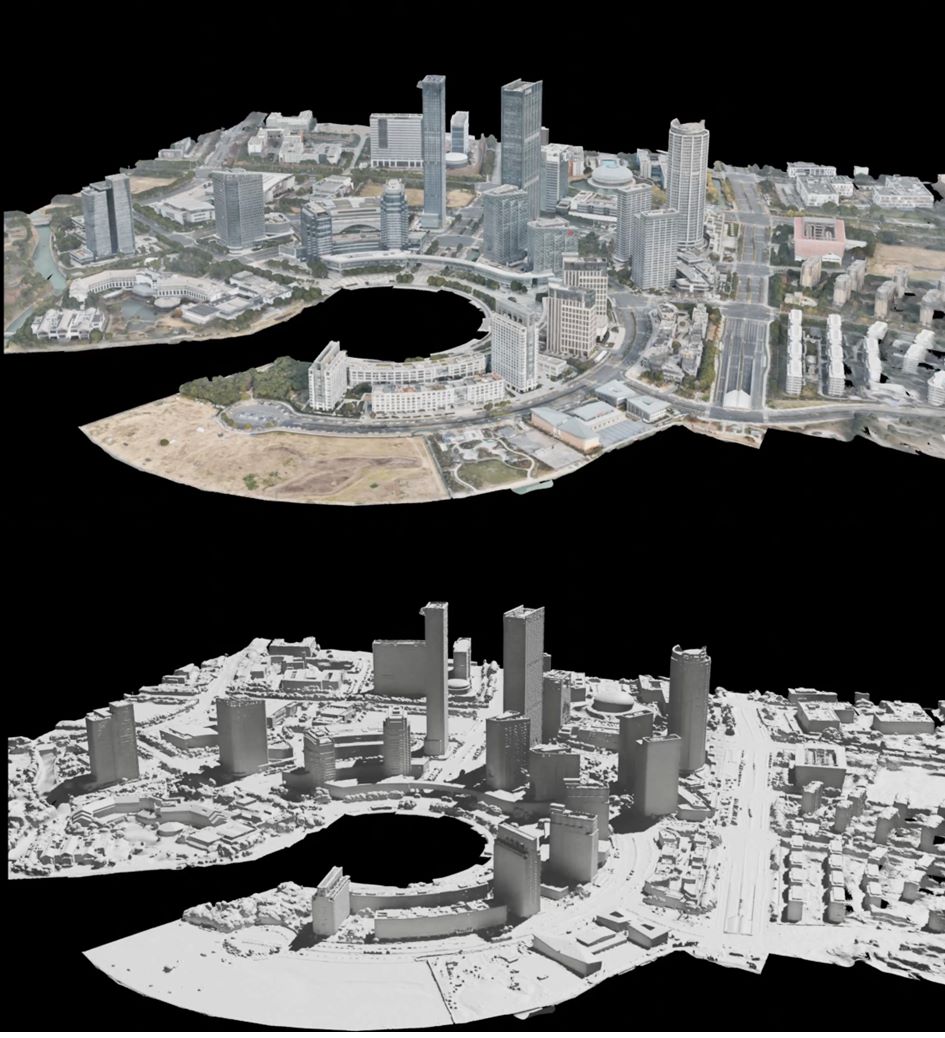}
    \caption{\textbf{Mesh Results on Scene 2.}}
    \label{fig:scane2}
\end{figure*}

\end{document}